\documentclass[10pt,journal,compsoc]{IEEEtran}

\usepackage{amsmath}
\usepackage{diagbox}
\usepackage{graphics}
\usepackage{subfigure}
\usepackage{epsfig}
\usepackage{threeparttable}
\usepackage{xspace}
\usepackage{siunitx}
\usepackage{graphicx}
\usepackage{amssymb}
\usepackage{threeparttable}
\usepackage{color}
\usepackage[normalem]{ulem}
\usepackage{multirow}
\usepackage{float}
\usepackage{amsfonts}
\usepackage{bm}
\usepackage{array}
\usepackage[table]{xcolor}
\usepackage{colortbl}
\usepackage{diagbox}
\usepackage{rotating}
\usepackage{booktabs}
\usepackage{overpic}
\usepackage{enumitem}
\usepackage{colortbl}
\usepackage{dblfloatfix}

\usepackage{bm}
\usepackage{multirow}
\usepackage[export]{adjustbox}

\usepackage{colortbl}
\definecolor{lightgray}{gray}{.9}
\definecolor{deepgray}{gray}{.8}

\usepackage{threeparttable}
\newcolumntype{I}{!{\vrule width 1pt}}
\makeatletter
\newcommand{\thickhline}{%
    \noalign {\ifnum 0=`}\fi \hrule height 1pt
    \futurelet \reserved@a \@xhline
}
\makeatother
\definecolor{mygray}{gray}{.9}
\usepackage{float}
\usepackage[ruled, vlined]{algorithm2e}
\usepackage{pifont}
\usepackage{ragged2e}

\usepackage[pagebackref=false,breaklinks=true,colorlinks,bookmarks=false]{hyperref}

\definecolor{mygray}{gray}{.9}
\definecolor{mygreen}{RGB}{93,173,85}
\definecolor{mywarning}{RGB}{233,144,61}
\definecolor{DarkRed}{RGB}{0,0,0}
\definecolor{azure}{rgb}{0.0, 0.5, 1.0}
\definecolor{gray}{rgb}{0.3, 0.3, 0.3}
\definecolor{DarkGreen}{RGB}{42,110,63}
\newcommand{\hlg}[1]{\textcolor{mygreen}{#1}}
 
\newcommand{\pub}[1]{{\color{gray}{\footnotesize{[{#1}]}}}}

\newcommand{\tmark}{\ding{51}} 
\usepackage[capitalize]{cleveref}
\crefname{section}{Sec.}{Secs.}
\crefname{table}{Tab.}{Tabs.}
\crefname{section}{§}{§§}

\makeatletter
\DeclareRobustCommand\onedot{\futurelet\@let@token\@onedot}
\def\@onedot{\ifx\@let@token.\else.\null\fi\xspace}

\def\eg{\emph{e.g}\onedot} 
\def\ie{\emph{i.e}\onedot}

\makeatletter
\newenvironment{fullitemize}
{
\begin{itemize}[leftmargin=*]
	\setlength{\itemsep}{3pt}
	\setlength{\parsep}{-5pt}
	\setlength{\parskip}{-3pt}
	\setlength{\leftmargin}{-10pt}
}
{
\end{itemize}
}
\makeatother

\begin{document}
\title{Federated Learning for Generalization, Robustness, Fairness: A Survey and Benchmark}
\author{Wenke Huang, Mang Ye,~\IEEEmembership{Senior Member,~IEEE}~, Zekun Shi, Guancheng Wan, He Li, Bo Du,~\IEEEmembership{Senior Member,~IEEE}, Qiang Yang,~\IEEEmembership{Fellow,~IEEE}

\IEEEcompsocitemizethanks{
\IEEEcompsocthanksitem 
Wenke Huang, Mang Ye, Zekun Shi, Guancheng Wan, He Li, Bo Du, are with the School of Computer Science, Wuhan University, Wuhan, China. \protect
E-mail:\{wenkehuang, yemang\}@whu.edu.cn
\IEEEcompsocthanksitem 
Qiang Yang is with the Department of Computer Science and Engineering, Hong Kong University of Science and Technology, Hong Kong, China. Email: qyang@cse.ust.hk.
}
}

\markboth{Federated Learning for Generalization, Robustness, Fairness: A Survey and Benchmark}%
{Shell \MakeLowercase{\textit{et al.}}: Bare Demo of IEEEtran.cls for Journals}

\IEEEtitleabstractindextext{
\begin{abstract}
Federated learning has emerged as a promising paradigm for privacy-preserving collaboration among different parties. 
Recently, with the popularity of federated learning, an inﬂux of approaches have delivered towards different realistic challenges. In this survey, we provide a systematic overview of the important and recent developments of research on federated learning.
Firstly, we introduce the study history and terminology definition of this area. Then, we comprehensively review three basic lines of research: generalization, robustness, and fairness, by introducing their respective background concepts, task settings, and main challenges. We also offer a detailed overview of representative literature on both methods and datasets. We further benchmark the reviewed methods
on several well-known datasets. Finally, we point out several open issues in this ﬁeld and suggest opportunities for further research. We also
provide a public website to continuously track developments in this fast advancing ﬁeld: \url{https://github.com/WenkeHuang/MarsFL}.

\end{abstract}
}

\maketitle
\IEEEdisplaynontitleabstractindextext
\IEEEpeerreviewmaketitle

\newcommand{\digits}{{Digits}}
\newcommand{\mnist}{{MNIST}}
\newcommand{\mnistabbrv}{{M}}
\newcommand{\usps}{{USPS}}
\newcommand{\uspsabbrv}{{U}}
\newcommand{\svhn}{{SVHN}}
\newcommand{\svhnabbrv}{{Sv}}
\newcommand{\syn}{{SYN}}
\newcommand{\synabbrv}{{Sy}}

\newcommand{\officecaltech}{{Office Caltech}}
\newcommand{\caltech}{{Caltech}}
\newcommand{\caltechabbrv}{{Ca}}
\newcommand{\amazon}{{Amazon}}
\newcommand{\amazonabbrv}{{Am}}
\newcommand{\webcam}{{Webcam}}
\newcommand{\webcamabbrv}{{W}}
\newcommand{\dslr}{{DSLR}}
\newcommand{\dslrabbrv}{{D}}

\newcommand{\caltechtfs}{{Caltech-256}}
\newcommand{\officeto}{{Office31}}

\newcommand{\officehome}{{Office-Home}}
\newcommand{\art}{{Art}}
\newcommand{\artabbrv}{{Ar}}
\newcommand{\clipart}{{Clipart}}
\newcommand{\clipartabbrv}{{Cl}}
\newcommand{\product}{{Product}}
\newcommand{\productabbrv}{{P}}
\newcommand{\realworld}{{Real World}}
\newcommand{\realworldabbrv}{{RW}}

\newcommand{\pacs}{{PACS}}
\newcommand{\photo}{{Photo}}
\newcommand{\photoabbrv}{{P}}
\newcommand{\artpainting}{{Art Painting}}
\newcommand{\artpaintingabbrv}{{AP}}
\newcommand{\cartoon}{{Cartoon}}
\newcommand{\cartoonabbrv}{{Ct}}
\newcommand{\sketch}{{Sketch}}
\newcommand{\sketchabbrv}{{Sk}}


\newcommand{\cifar}{{Cifar}}
\newcommand{\weperson}{{WePerson}}
\newcommand{\imagenet}{{ImageNet}}
\newcommand{\cifarhun}{{Cifar-100}}
\newcommand{\cifarten}{{Cifar-10}}
\newcommand{\tyimg}{{Tiny-ImageNet}}
\newcommand{\coco}{{COCO}}
\newcommand{\market}{{Market1501}}
\newcommand{\fashionmnist}{{Fashion-MNIST}}

\newcommand{\simplecnn}{{SimpleCNN}}
\newcommand{\resnet}{{ResNet}}
\newcommand{\resnetten}{{ResNet-10}}
\newcommand{\resnettwelve}{{ResNet-12}}
\newcommand{\resneteighteen}{{ResNet-18}}
\newcommand{\resnettwenty}{{ResNet-20}}
\newcommand{\resnetthirtyfour}{{ResNet-34}}
\newcommand{\resnetfifty}{{ResNet-50}}
\newcommand{\resnetonezeroone}{{ResNet-101}}
\newcommand{\resnetonefivetwo}{{ResNet-152}}

\newcommand{\resnext}{{ResNeXt}}
\newcommand{\ResNextFourEleven}{{ResNeXt4-11}}
\newcommand{\mobilenet}{{MobileNet}}
\newcommand{\efficientnet}{{EfficientNet}}
\newcommand{\densenet}{{DenseNet}}
\newcommand{\convnet}{{Convnet}}
\newcommand{\googlenet}{{GoogLeNet}}
\newcommand{\fbnet}{{Fbnet}}

\newcommand{\sgd}{{SGD}}
\newcommand{\adam}{{Adam}}

\newcommand{\finch}{{Finch}}
\newcommand{\hac}{{HAC}}
\newcommand{\kmeans}{{Kmeans}}
\newcommand{\cka}{{CKA}}
\newcommand{\tsne}{{t-SNE}}
\newcommand{\dbscan}{{DBSCAN}}

\newcommand{\ce}{{CE Loss}}
\newcommand{\kl}{{KL Loss}}
\newcommand{\triplet}{{Triplet Loss}}
\newcommand{\normface}{{Normface}}
\newcommand{\sphereface}{{Sphereface}}
\newcommand{\rce}{{RCE}}
\newcommand{\proxyanchor}{{Proxy-Anchor}}
\newcommand{\atk}{{$AT_k$}}
\newcommand{\centerloss}{{Center}}
\newcommand{\largemargin}{{L-Softmax}}
\newcommand{\arcface}{{ArcFace}}
\newcommand{\polyloss}{{PolyLoss}}

\newcommand{\fedavg}{{FedAvg}}
\newcommand{\fedprox}{{FedProx}}
\newcommand{\fedcurv}{{FedCurv}}
\newcommand{\scaffold}{{SCAFFOLD}}
\newcommand{\feddyn}{FedDyn}
\newcommand{\moon}{{MOON}}
\newcommand{\fedproc}{{FedProc}}
\newcommand{\ccvr}{{CCVR}}
\newcommand{\fedaux}{{FEDAUX}}
\newcommand{\fedmgda}{{FedMGDA+}}
\newcommand{\sparsefed}{{SparseFed}}
\newcommand{\feddc}{{FedDC}}
\newcommand{\pfedme}{{pFedME}}
\newcommand{\dqs}{{DQS}}
\newcommand{\trimmedmedian}{{Trim Median}}
\newcommand{\favor}{{FAVOR}}
\newcommand{\fedproto}{{FedProto}}
\newcommand{\rlr}{{RLR}}
\newcommand{\mccda}{{MCC-DA}}
\newcommand{\provablefl}{{ProvableFL}}
\newcommand{\fedagg}{{FedAgg}}
\newcommand{\feddg}{{FedDG}}
\newcommand{\cerp}{{CerP}}
\newcommand{\ida}{{IDA}}
\newcommand{\dpa}{{DPA}}
\newcommand{\ircmsda}{{IRCMSDA}}
\newcommand{\creff}{{CReFF}}
\newcommand{\fedada}{{FedAda}}
\newcommand{\fedsae}{{FedSAE}}
\newcommand{\autofedavg}{{Auto-FedAvg}}
\newcommand{\fedbn}{{FedBN}}
\newcommand{\astraea}{{Astraea}}
\newcommand{\fedacs}{{FedACS}}
\newcommand{\dcadam}{{DC-Adam}}
\newcommand{\aegr}{{AEGR}}
\newcommand{\comda}{{Co-MDA}}
\newcommand{\divfl}{{DivFL}}
\newcommand{\fedss}{{FEDSS}}
\newcommand{\climb}{{CLIMB}}
\newcommand{\fedci}{{FedCI}}
\newcommand{\fccl}{{FCCL}}
\newcommand{\mincost}{{MinCost}}
\newcommand{\fedcg}{{FedCG}}
\newcommand{\dsfl}{{DS-FL}}
\newcommand{\fedlaw}{{FEDLAW}}
\newcommand{\bnpfl}{{BNPFL}}
\newcommand{\fedcams}{{FedCAMS}}
\newcommand{\xormixup}{{XorMixUp}}
\newcommand{\fedspeed}{{FEDSPEED}}
\newcommand{\fada}{{FADA}}
\newcommand{\fedfv}{{FedFV}}
\newcommand{\fedasync}{{FedAsync}}
\newcommand{\fedssl}{{FedSSL}}
\newcommand{\furl}{{FURL}}
\newcommand{\fedmd}{{FedMD}}
\newcommand{\cronus}{{Cronus}}
\newcommand{\fedmdnfdp}{FEDMD-NFDP}
\newcommand{\feddf}{{FedDF}}
\newcommand{\fedrad}{{FedRAD}}
\newcommand{\rhfl}{{RHFL}}
\newcommand{\afl}{{AFL}}
\newcommand{\lgfedavg}{LG-FEDAVG}
\newcommand{\fedrs}{{FedRS}}
\newcommand{\adcol}{{ADCOL}}
\newcommand{\fedmix}{{FedMix}}
\newcommand{\fedgkt}{{FedGKT}}
\newcommand{\flmoe}{{FLMoE}}
\newcommand{\fedopt}{{FedOPT}}
\newcommand{\fml}{{FML}}
\newcommand{\fedufo}{{FedUFO}}
\newcommand{\fedsam}{{FedSAM}}
\newcommand{\knnper}{{kNN-Per}}
\newcommand{\fedlc}{{FedLC}}
\newcommand{\fednova}{{FedNova}}
\newcommand{\pfedla}{{pFedLA}}
\newcommand{\orchestra}{{Orchestra}}
\newcommand{\fedrod}{{FED-ROD}}
\newcommand{\fedsm}{{FedSM}}
\newcommand{\fedmlb}{{FedMLB}}
\newcommand{\fedhenn}{{FedHeNN}}
\newcommand{\fedtwo}{{Fed$^2$}}
\newcommand{\fade}{{FADE}}
\newcommand{\mocha}{{MOCHA}}
\newcommand{\fedpara}{{FedPara}}
\newcommand{\waffle}{{WAFFLe}}
\newcommand{\matcha}{{Matcha}}
\newcommand{\agnosticfl}{{AFL}}
\newcommand{\fedper}{{FEDPER}}
\newcommand{\comt}{{CoMT}}
\newcommand{\fpca}{{FPCA}}
\newcommand{\lotteryfl}{{LotteryFL}}
\newcommand{\rcfl}{{RCFL}}
\newcommand{\fsmafl}{{FSMAFL}}
\newcommand{\fedreid}{{FedReID}}
\newcommand{\fedamp}{{FedAMP}}
\newcommand{\pfedhn}{{pFedHN}}
\newcommand{\fedmatch}{{FedMatch}}
\newcommand{\fedbe}{{FEDBE}}
\newcommand{\ditto}{{Ditto}}
\newcommand{\fedrecon}{{FEDRECON}}
\newcommand{\fedgen}{{FEDGEN}}
\newcommand{\fedcgan}{{FedCGAN}}
\newcommand{\fedad}{{FedAD}}
\newcommand{\fedzdac}{{Fed-ZDAC}}
\newcommand{\heterofl}{{HeteroFL}}
\newcommand{\fedfomo}{{FedFomo}}
\newcommand{\soteria}{{Soteria}}
\newcommand{\fedagm}{{FedAGM}}
\newcommand{\fednew}{{FedNew}}
\newcommand{\fedalign}{{FedAlign}}
\newcommand{\feddst}{{FedDST}}
\newcommand{\safe}{{Safe}}
\newcommand{\rrfl}{{RRFL}}
\newcommand{\fedcorr}{{FedCorr}}
\newcommand{\splitmix}{{SplitMix}}
\newcommand{\fedcor}{{FedCor}}
\newcommand{\fedftg}{{FedFTG}}
\newcommand{\fedbabu}{{FedBABU}}
\newcommand{\sfl}{{SFL}}
\newcommand{\gifair}{{GIFAIR}}
\newcommand{\fedsoft}{{FedSoft}}
\newcommand{\fedreg}{{FedReg}}
\newcommand{\fedmat}{{FedMAT}}
\newcommand{\mabrfl}{{MAB-RFL}}
\newcommand{\cfed}{{CFeD}}
\newcommand{\cgpfl}{{CGPFL}}
\newcommand{\maker}{{MaKEr}}
\newcommand{\harmofl}{{HarmoFL}}
\newcommand{\fedpa}{{FEDPA}}
\newcommand{\spahm}{{SPAHM}}
\newcommand{\fedma}{{FedMA}}
\newcommand{\fedntd}{{FedNTD}}
\newcommand{\feddualavg}{{FEDDUALAVG}}
\newcommand{\fedrn}{{FedRN}}
\newcommand{\fedpcl}{{FedPCL}}
\newcommand{\pgfl}{{PGFL}}
\newcommand{\ifca}{IFCA}
\newcommand{\ktpfl}{{KT-pFL}}
\newcommand{\fedscale}{{FedScale}}
\newcommand{\fedfa}{{FedFA}}
\newcommand{\fpl}{{FPL}}
\newcommand{\flhc}{{FL+HC}}
\newcommand{\faug}{{FAug}}
\newcommand{\cfl}{{CFL}}
\newcommand{\mcfl}{{MFCL}}
\newcommand{\multikrum}{{Multi Krum}}
\newcommand{\fedgroup}{{FedGroup}}
\newcommand{\hypcluster}{{HYPCLUSTER}}
\newcommand{\fedpr}{{FedPR}}
\newcommand{\dynafed}{{DYNAFED}}
\newcommand{\fedce}{{FedCE}}
\newcommand{\qffl}{{qFFL}}
\newcommand{\fedfusion}{{FedFusion}}
\newcommand{\dafkd}{{DaFKD}}
\newcommand{\fedpvr}{{FedPVR}}
\newcommand{\feddm}{{FedDM}}
\newcommand{\ot}{{OT}}
\newcommand{\fedsim}{{FedSim}}
\newcommand{\crfl}{{CRFL}}
\newcommand{\smartfl}{{SmartFL}}
\newcommand{\foolsgold}{{FoolsGold}}
\newcommand{\krum}{{Krum}}
\newcommand{\bulyan}{{Bulyan}}
\newcommand{\rfa}{{RFA}}
\newcommand{\rsa}{{RSA}}
\newcommand{\dncagg}{{{DnC}}}
\newcommand{\median}{{Median}}
\newcommand{\afa}{{AFA}}
\newcommand{\aaa}{{AAA}}
\newcommand{\faba}{{FABA}}
\newcommand{\fltrust}{{FLTrust}}
\newcommand{\sageflow}{{Sageflow}}
\newcommand{\dba}{{DBA}}
\newcommand{\rbtm}{{RB-TM}}
\newcommand{\ipm}{{IPM}}
\newcommand{\lie}{{Little Is Enough}}
\newcommand{\creamfl}{{CreamFL}}
\newcommand{\fedseg}{{FedSeg}}
\newcommand{\dimkrum}{{Dim-Krum}}
\newcommand{\fedsi}{{FedSI}}
\newcommand{\jupiter}{{Jupiter}}
\newcommand{\fedtfi}{{FedTFI}}
\newcommand{\fedbeal}{{FedBEAL}}
\newcommand{\fedspace}{{FedSpace}}
\newcommand{\cbafed}{{CBAFed}}
\newcommand{\fedbr}{{FedBR}}
\newcommand{\copa}{{COPA}}
\newcommand{\fedsr}{{FedSR}}
\newcommand{\gradma}{{GradMA}}
\newcommand{\fedga}{{FedGA}}
\newcommand{\fednoro}{{FedNoRo}}
\newcommand{\fedlsr}{{FedLSR}}
\newcommand{\ccst}{{CCST}}
\newcommand{\iopfl}{{IOP-FL}}
\newcommand{\csac}{{CSAC}}
\newcommand{\mckd}{{MCKD}}
\newcommand{\fedinb}{{FedINB}}
\newcommand{\cffl}{{CFFL}}
\newcommand{\fcfl}{{FCFL}}
\newcommand{\fedfaim}{{FedFAIM}}
\newcommand{\cgsv}{{CGSV}}
\newcommand{\apple}{{APPLE}}
\newcommand{\fedthe}{{FedTHE}}
\newcommand{\fedclip}{{FedCLIP}}
\newcommand{\feddecorr}{{FedDecorr}}
\newcommand{\flip}{{FLIP}}
\newcommand{\depthfl}{{DepthFL}}
\newcommand{\fedpac}{{FedPAC}}
\newcommand{\ilrg}{{iLRG}}
\newcommand{\feddar}{{FedDAR}}
\newcommand{\fedhkd}{{FedHKD}}
\newcommand{\batfl}{{BatFL}}
\newcommand{\kdthreea}{{KD3A}}
\newcommand{\fedadg}{{FedADG}}
\newcommand{\fedadgiotj}{{FedADG}}
\newcommand{\fthreeba}{{F3BA}}
\newcommand{\dbfat}{{DBFAT}}
\newcommand{\fedala}{{FedALA}}
\newcommand{\defl}{{DeFL}}
\newcommand{\fedmdfg}{{FedMDFG}}
\newcommand{\fednh}{{FedNH}}
\newcommand{\fedrbn}{{FedBRN}}
\newcommand{\fairfed}{{FairFed}}
\newcommand{\clusterattack}{{ClusterAttack}}
\newcommand{\fang}{{Fang}}

\newcommand{\hfl}{{Horizontal Federated Learning}}
\newcommand{\hflabbrv}{{HFL}}
\newcommand{\vfl}{{Vertical Federated Learning}}
\newcommand{\vflabbrv}{{VFL}}
\newcommand{\ftl}{{Federated Transfer Learning}}
\newcommand{\ftlabbrv}{{FTL}}
\newcommand{\llm}{{Large Language Model}}
\newcommand{\llmabbrv}{{LLM}}

\newcommand{\gfl}{{Generalizable Federated Learning}}
\newcommand{\gflabbrv}{{GFL}}
\newcommand{\crosscal}{{Cross Calibration}}
\newcommand{\crosscalabbrv}{{CorrCal}}
\newcommand{\clientreg}{{Client Regularization}}
\newcommand{\clientregabbrv}{{CliReg}}
\newcommand{\clientaug}{{Client Augmentation}}
\newcommand{\clientaugabbrv}{{CliAug}}
\newcommand{\serverope}{{Server Operation}}
\newcommand{\serveropeabbrv}{{SerOpe}}
\newcommand{\optimcal}{{Client Regularization}}
\newcommand{\optimcalabbrv}{{OptCal}}
\newcommand{\ungene}{{Unknown Generalization}}
\newcommand{\ungeneabbrv}{{UnkGen}}
\newcommand{\fda}{{Federated Domain Adaptation}}
\newcommand{\fdaabbrv}{{FDA}}
\newcommand{\fdg}{{Federated Domain Generalization}}
\newcommand{\fdgabbrv}{{FDG}}
\newcommand{\croclishift}{{Cross-Client Shift}}
\newcommand{\outclishift}{{Out-Client Shift}}

\newcommand{\rfl}{{Robust Federated Learning}}
\newcommand{\rflabbrv}{{RFL}}
\newcommand{\byzantoler}{{Byzantine Tolerance}}
\newcommand{\backdefen}{{Backdoor Defense}}
\newcommand{\distoler}{{Distance Base Tolerance}}
\newcommand{\statoler}{{Statistics Distribution Tolerance}}
\newcommand{\protoler}{{Proxy Dataset Tolerance}}

\newcommand{\refdefen}{{Model Refinement Defense}}
\newcommand{\aggdefen}{{Robust Aggregation Defense}}
\newcommand{\cerdefen}{{Certified Robustness Defense}}

\newcommand{\ffl}{{Fair Federated Learning}}
\newcommand{\fflabbrv}{{FFL}}
\newcommand{\perfair}{{Performance Fairness}}
\newcommand{\perdebioptim}{{Performance Debias Optimization}}
\newcommand{\perfairrewei}{{Performance Debias Rewighting}}
\newcommand{\colfair}{{Collaboration Fairness}}
\newcommand{\indveval}{{Individual Contribution Evaluation}}
\newcommand{\margeval}{{Marginal Contribution Evaluation}}

\newcommand{\poiatt}{{Poisoning Attack}}
\newcommand{\bayatt}{{Byzantine Attack}}
\newcommand{\databayatt}{{Data-Based Byzantine Attack}}
\newcommand{\modelbayatt}{{Model-Based Byzantine Attack}}
\newcommand{\bacatt}{{Backdoor Attack}}
\newcommand{\priinf}{{Privacy Inference}}
\newcommand{\predbias}{{Prediction Biases}}
\newcommand{\rewaconf}{{Reward Conflict}}

\newcommand{\labelshift}{{Label Skew}}
\newcommand{\domainshift}{{Domain Skew}}

\newcommand{\pairflip}{{Pair Flipping}}
\newcommand{\pairflipabbrv}{{PairF}}
\newcommand{\symflip}{{Symmetry Flipping}}
\newcommand{\symflipabbrv}{{SymF}}
\newcommand{\randomnoise}{{Random Noise}}
\newcommand{\randomnoiseabbrv}{{RanN}}
\newcommand{\addnoise}{{Add Noise}}
\newcommand{\addnoiseabbrv}{{AddN}}
\newcommand{\minmax}{{Min-Max}}
\newcommand{\minmaxabbrv}{{MiMa}}
\newcommand{\minsum}{{Min-Sum}}
\newcommand{\minsumabbrv}{{MiSu}}
\newcommand{\lieabbrv}{{LIE}}

\newcommand{\backdoorabbrv}{{Bac}}
\newcommand{\semanticbackdoorabbrv}{{Sem Bac}}

\newcommand{\croclienacc}{{Cross-Client Accuracy}}
\newcommand{\outclienacc}{{Out-Client Accuracy}}
\newcommand{\accdeclineimp}{{Accuracy Decline Impact}}
\newcommand{\attsuccerat}{{Attack Success Rate}}
\newcommand{\perfovderv}{{Performance Deviation}}
\newcommand{\conmatchdeg}{{Contribution Match Degree}}

\IEEEraisesectionheading{
\section{Introduction}
\label{sec:introduction}
}

\IEEEPARstart{C}{urrent} development of deep learning has caused significant changes in numerous research fields, and had profound impacts on every nook and cranny of societal and industrial sectors, including computer vision \cite{ImageNet_CVPR09,ImageNet_IJCV15,ViT_ICLR21,ResNet_CVPR16,ResNeXt_CVPR17,DenseNet_PAMI19,GenPretrainedTrans_arXiv23}, natural language processing \cite{Attention_NeurIPS17,BERT_arXiv18}, multi-modal learning {\cite{CLIP_ICML21,DeCLIP_ICLR22,CaFo_CVPR23}}, medical analysis {\cite{HotProtein_ICLR23}}, and more. However, the success of deep learning heavily relies on large-scale data and there has been increasing awareness in the public and scientific communities for data privacy. Specifically, in the real world, data is commonly distributed among different entities (\eg, edge devices and companies). Due to the increasing emphasis on data sensitivity, strict legislations {\cite{IPR_06,ToFairnessinDataSharing_NEJM16,GDPR17,CCPA_18}} have been proposed to govern data collection and utilization. Thus, the traditional centralized training paradigm, which requires to aggregate data, fails to deploy in the practical setting. 
Driven by such realistic challenges, federated learning (FL) {\cite{FLStrategies_arXiv16,FLOptimization_arXiv16,FedAvg_AISTATS17,FederatedMLConApp_TIST19,LAQ_PAMI20,OFL_PAMI21,BrainIoT_IOTJ21}} has emerged as a popular research field because it can train a global model for different participants without centralizing data owned by the distributed parties. Roughly speaking, the classical federated paradigm can be abstractly divided into the following two steps: server-side collaboration and client-side optimization. The former could be regarded as that a central server aggregates parameters from participants and then distributes the global model (averaged parameters) back. The latter represents that the client optimizes the distributed model on the local private data. Therefore, FL achieves the privacy-preserving collaboration to learn a shared model without data consolidation. Despite great advancements in federated learning \cite{FLKeyboard_arXiv18,FLStroke_arXiv20,FedReID_MM20,FLHumorReco_IJCAI2020,Fedvision_AAAI20,FedReID_AAAI21,FLbigdataforRareCaner_NatCom22,MSDformer_TGRS23}, current federation has three major challenges as the following:

\noindent$\bullet$~\textbf{Generalization}. The distributed data is normally collected from different sources with diverse preferences and naturally brings the non-independent and identically distributed (Non-IID) characteristics \cite{FedCurv_NeurIPSW19,CBFL_JBI19,NonIIDQuagmireofFL_ICML20,FedProx_MLSys2020,FedBN_ICLR21,FedPerGNN_NC22,FedStar_AAAI23}. Owing to this distribution discrepancy, there exists two major distribution shift types.  
\noindent \textbf{i)} \croclishift{}:  Distributed data from different clients is inevitable to appear a different
underlying distribution, incurring large heterogeneity among client data. Under this circumstance, each client optimizes toward its distinct local empirical minimum. resulting in divergent local optimization directions \cite{MOON_CVPR21,CCVR_NeurIPS21,FedUFO_ICCV21,FedGA_AAAI22,FedSAM_ICML22}. Thus, the federation presents slow convergence and unsatisfying performance. 
\noindent \textbf{ii)} \outclishift{}:
The federated system only incorporates knowledge from participating clients and optimizes toward fitting online client distribution. Therefore, it would be vulnerable under the \outclishift{} \cite{GeneralizationinFL_ICLR22,FADA_ICLR20,FedDG_CVPR21,IOPFL_TMI23,FedTHE_ICLR23}. Specifically, when deploying the federated model to outer client distribution (\ie, unseen testing domains), the federation fails to adapt to the target distribution and inevitable performance degradation is commonly observed.
All in all, achieving a generalizable federated model acts as a foundational requirement to boost the convergence speed and performance accuracy for the real-world data distribution.

\noindent$\bullet$~\textbf{Robustness}. As a privacy-aware collaborative paradigm, ensuring federated robustness plays a crucial role in guaranteeing federated effectiveness. In practice, due to its distributed nature, federated learning fails to ensure the client trustworthiness and is highly vulnerable to different malicious behaviors. A small set of malicious clients can easily manipulate the training process by uploading poisoned local models to the server
\noindent \textbf{i)} \bayatt{}: malicious clients endeavor to falsify real network updates to attack the federated convergence and performance,  {\cite{AdvML_SAI11,PoisoningattacksSVM_ICML12,Badnets_arXiv17,Bulyan_ICML18}}. Existing attacks mainly focus on  poisoning client local training data, \ie, \databayatt{} {\cite{SymmetryFlip_NeurIPS15,PairFlip_NeurIPS18}} and directly manipulating the model parameters, \modelbayatt{} {\cite{LIE_NeurIPS19,IPM_UAI20,Fang_USENIX20}}
{\color{DarkRed}
\noindent \textbf{i)} \bacatt{}: evils inject the backdoor trigger to induce the global model to learn misleading information, so as to predict attacker-chosen labels for adversarial examples while maintaining stable performance on the main task {\cite{TargetBackdoor_arXiv17,BackdoorviaInvisiblePerturbation_arXiv18,BadNets_Access19,HowtoBackDoorFL_AISTATS20}}. Furthermore, backdoor attacks utilize the distributed paradigm to consider more stealthy trigger patterns via decomposing the global trigger pattern into separate local patterns to easily dodge various defensive mechanisms while successfully launching distributed backdoor attacks \cite{DBA_ICLR20,CerP_AAAI23}.}

\noindent$\bullet$~\textbf{Fairness}. Federated learning functions as a collaborative paradigm. The crucial cooperation pre-requirement is to satisfy the multi-party interest allocation for its sustained development viability {\cite{FairandPriFedModel_TPDS20,FairandPrivacyFL_arXiv23,FAFL_TNNLS23}}. Within this context, two primary interest conflicts groups emerge out. 
\noindent \textbf{i)} \colfair{}: in the absence of equitable credit assignment and just reward allocation, clients lack the incentive to actively participate in the federated collaboration \cite{FedCI_BigData19,IncentiveFL_TOJ19,CFFL_FL20}. 
\noindent \textbf{ii)} \perfair{}: the federated model may result in biased predictions, thereby yielding sub-optimal performance for clients characterized by limited data scale or distinct data distributions from the rest \cite{AFL_ICML19,qFFL_ICLR20,FADE_SIGKDD21}.
Consequently, the maintenance of both collaboration fairness and performance fairness significantly determines whether federated learning can function in a robust and enduring manner.
\noindent{} These shortcomings severely hinder the adoption of federated learning in realistic applications, such as medical diagnosis {\cite{FL4SmartHealth_CSUR22}}, autonomous driving {\cite{FLforAutoDriv_IV22}}, and market finance analysis {\cite{FMLforFraudulentCreditCard_IJCAI21}}, which lead to the increasing concern on the contemporary federated learning techniques.

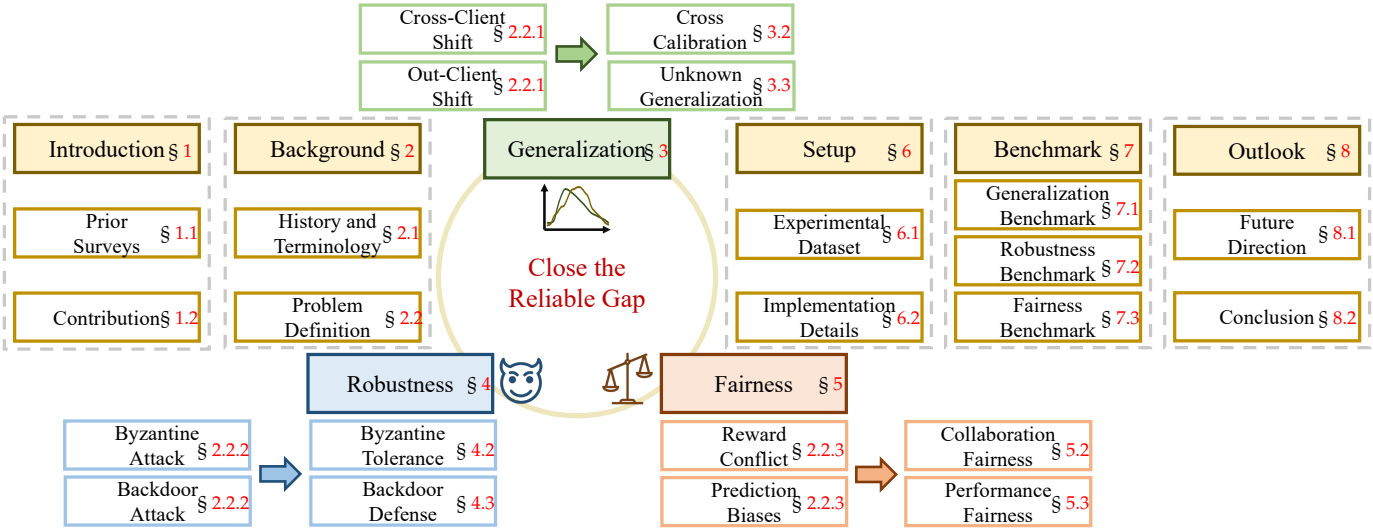
\begin{figure*}[t]
\centering
\includegraphics[width=\linewidth]{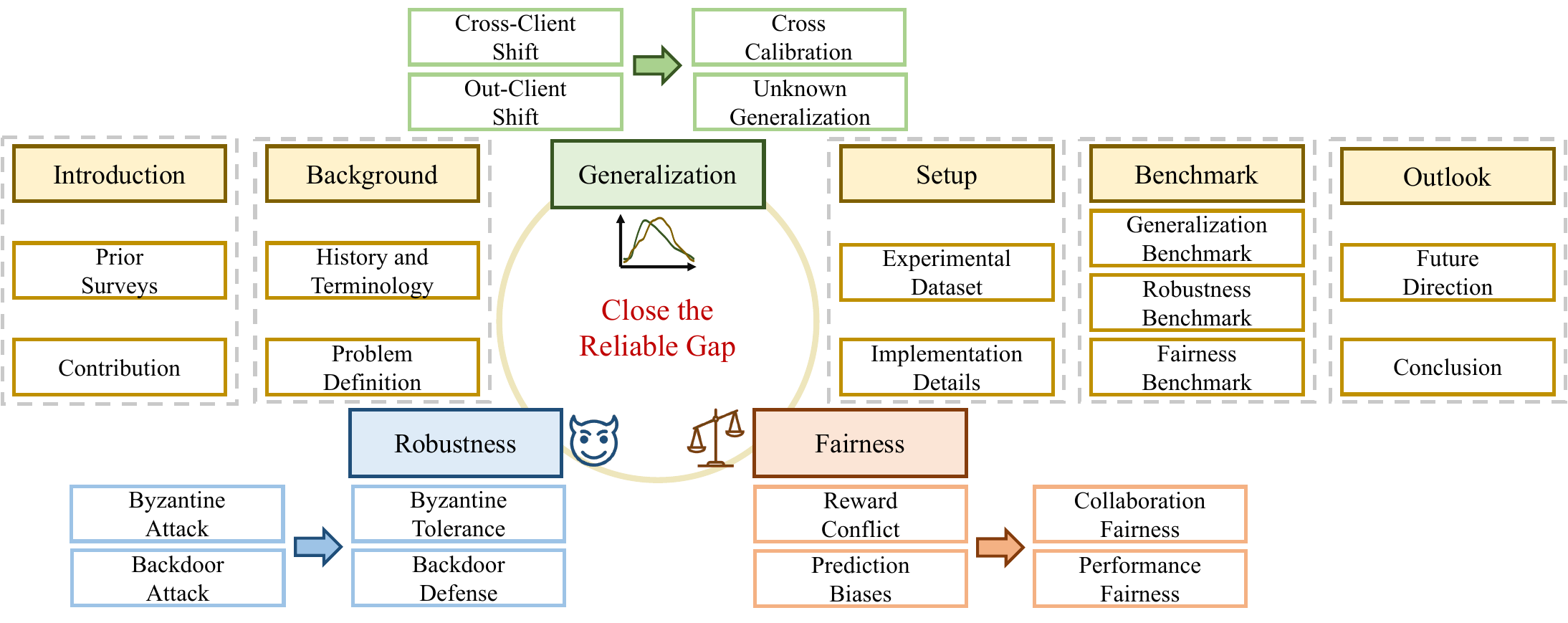}
\put(-455,145){\scriptsize{\cref{sec:introduction}}}
\put(-458,114){\scriptsize{\cref{sec:priorsurveys}}}
\put(-458,82){\scriptsize{\cref{sec:contribution}}}
\put(-371,145){\scriptsize{\cref{sec:background}}}
\put(-374,114){\scriptsize{\cref{sec:historyandteriminology}}}
\put(-373,82){\scriptsize{\cref{sec:prob}}}
\put(-183,145){\scriptsize{\cref{sec:setup}}}
\put(-185,114){\scriptsize{\cref{sec:datasets}}}
\put(-185,82){\scriptsize{\cref{sec:implement_details}}}
\put(-99,145){\scriptsize{\cref{sec:benchmark}}}
\put(-102,123){\scriptsize{\cref{sec:gen_compare}}}
\put(-102,101){\scriptsize{\cref{sec:robust_compare}}}
\put(-102,82){\scriptsize{\cref{sec:fairness_compare}}}
\put(-17,145){\scriptsize{\cref{sec:outlook}}}
\put(-20,114){\scriptsize{\cref{sec:future_direction}}}
\put(-20,82){\scriptsize{\cref{sec:conclusion}}}
\put(-342,57){\scriptsize{\cref{sec:rfl}}}
\put(-346,32){\scriptsize{\cref{sec:byzantoler}}}
\put(-346,12){\scriptsize{\cref{sec:backdefen}}}
\put(-444,32){\scriptsize{\cref{sec:maliciousbehavior}}}
\put(-444,12){\scriptsize{\cref{sec:maliciousbehavior}}}
\put(-209,57){\scriptsize{\cref{sec:ffl}}}
\put(-219,32){\scriptsize{\cref{sec:interestconflict}}}
\put(-219,12){\scriptsize{\cref{sec:interestconflict}}}
\put(-121,32){\scriptsize{\cref{sec:colfair}}}
\put(-121,12){\scriptsize{\cref{sec:perfair}}}
\put(-275,145){\scriptsize{\cref{sec:gfl}}}
\put(-332,190){\scriptsize{\cref{sec:distributedshift}}}
\put(-332,170){\scriptsize{\cref{sec:distributedshift}}}
\put(-234,190){\scriptsize{\cref{sec:crosscal}}}
\put(-234,170){\scriptsize{\cref{sec:ungene}}}
\vspace{-10pt}
\caption{
\textbf{Overview of the survey}. Best viewed in color.
}  
\label{fig:framework}
\vspace{-10pt}
\end{figure*}

\begin{table}[t]\small
\caption{{
\textbf{Overview of related surveys}. See details in \cref{sec:priorsurveys}.
}
}
\label{tab:ourscomputation}
\vspace{-10pt}
\centering
{
\resizebox{\columnwidth}{!}{
\setlength\tabcolsep{1pt}
\renewcommand\arraystretch{1.2}
\begin{tabular}{rIcccc}
\hline\thickhline
\rowcolor{mygray}
 Work & Generalization & Robustness & Fairness & Benchmark \\
\hline\hline 
\pub{arXiv'18} \cite{FLwithNonIID_arXiv18} & Cross-Client Shift & & & \\
\pub{TIST'19} \cite{FederatedMLConApp_TIST19} & Cross-Client Shift &  & & \\
\pub{WS4'20} \cite{SurPesoFL_WorldS420} & Cross-Client Shift & & & \\ 
\pub{SPM'20}\cite{FLChallengesMethodsDirection_SPM20}& Cross-Client Shift &  Byzantine  & & \\ 
\pub{arXiv'20}\cite{ThreatstoFL_arXiv20} &  & \hlg{\tmark{}} & & \\
\pub{NC'21}\cite{FLonNONIDD_NC21} & Cross-Client Shift & & & \\
\pub{FTML'21}\cite{AdvancesandProblemFL_FTML21}& Cross-Client Shift & \hlg{\tmark{}} & \hlg{\tmark{}} & \\
\pub{FGCS'21}\cite{SurveyonSecurityandPrivacyofFL_FGCS21} & Cross-Client Shift & \hlg{\tmark{}} &  & \\
\pub{TKDE'21}\cite{SurveyonFLSystem_TKDE21} &  Cross-Client Shift & \hlg{\tmark{}} & Performance & \\
\pub{arXiv'21}\cite{AsyFL_arXiv21} &  Cross-Client Shift & \hlg{\tmark{}} &  & \\
\pub{TrustCom'22} \cite{ByzantineAttacksinFL_TrustCom22} &  & Byzantine  & &  \hlg{\tmark{}}\\
\pub{TKDE'22} \cite{FLONNonIID_TKDE22} &  Cross-Client Shift & & &  \hlg{\tmark{}}\\
\pub{CSUR'22} \cite{FL4SmartHealth_CSUR22} &  Cross-Client Shift & & & \\
\pub{arXiv'22} \cite{UniFed_arXiv22} &  Cross-Client Shift & & &  \hlg{\tmark{}} \\
\pub{TNNLS'22} \cite{PrivacyrobustnessFL_TNNLS22} & &  \hlg{\tmark{}} &  & \\
\pub{FGCS'22} \cite{NonIIDFL_FGCS22} & Cross-Client Shift & & & \\
\pub{arXiv'23} \cite{FairandPrivacyFL_arXiv23} & &  &  \hlg{\tmark{}} & \\
\pub{arXiv'23} \cite{FedDGSurvey_arXiv23} & Out-Client Shift & & & \\
\pub{ariv'23} \cite{SurveywhattoshareinFL_arXiv23} &  Cross-Client Shift &  & & \hlg{\tmark{}}\\
\pub{TNNLS'23} \cite{FAFL_TNNLS23} &  & & \hlg{\tmark{}}\\ 
\pub{CSUR'23} \cite{HFL_CSUR23} &  Cross-Client Shift & & & \\
\hline \hline
Ours & \hlg{\tmark{}} & \hlg{\tmark{}} &  \hlg{\tmark{}} & \hlg{\tmark{}}\\
\end{tabular}}}
\end{table}

\subsection{Prior Surveys}
\label{sec:priorsurveys}
As FL research has become a prominent research field in recent years, a large amount federated survey papers have emerged. Pioneering works \cite{FederatedMLConApp_TIST19,FLChallengesMethodsDirection_SPM20,FederatedML_IEEECST21,SurveyonFLSystem_TKDE21,AdvancesandProblemFL_FTML21,FLThreat_IF23} basically focus on the conceptual framework and macro guidance and neglect in-depth exploration of specific challenges and problems. 
The majority focuses on the data heterogeneity problem {\cite{FLwithNonIID_arXiv18,SurPesoFL_WorldS420,AsyFL_arXiv21,FLonNONIDD_NC21,FLonNonIID_ExperimentalStudy_ICDE22,FLONNonIID_TKDE22,FedDGSurvey_arXiv23,HFL_CSUR23,FLwoFUllLabels_arXiv23,UniFed_arXiv22}} from different angles. For example {\cite{FLwithNonIID_arXiv18,FLONNonIID_TKDE22,HFL_CSUR23}} tackle the non-IID issue in FL and provide detailed discussion on different specific data heterogeneity forms with respective federated solutions.
A few attempts respectively investigate the robustness \cite{FederatedMLConApp_TIST19,FLChallengesMethodsDirection_SPM20,ThreatstoFL_arXiv20,SurveyonSecurityandPrivacyofFL_FGCS21,ByzantineAttacksinFL_TrustCom22,SurveywhattoshareinFL_arXiv23,FairandPrivacyFL_arXiv23} and fairness \cite{FAFL_TNNLS23,SurveyonFLSystem_TKDE21} issues. 
Some surveys discuss specific real-world applications such as medical contexts \cite{FLinMedicalApp_FDSE21} and smart healthcare \cite{FL4SmartHealth_CSUR22} but lack the universality for different scenarios. For instance, \cite{FL4SmartHealth_CSUR22} concentrates on the federated healthcare topic, which mainly discusses scalable smart healthcare networks and applications, including healthcare management, remote health monitoring, and so on.  However, over-focusing on specific areas weakens their universality in other federated scenarios.
Importantly, although there are few surveys such as \cite{UniFed_arXiv22,FLONNonIID_TKDE22,ByzantineAttacksinFL_TrustCom22,SurveywhattoshareinFL_arXiv23} with empirical experiments analysis, their framework appears a strong entanglement with the specific problems and thus lacks the high flexibility towards realistic federated problems. Besides, we provide fruitful SOTA methods implements, which largely benefit relative researchers.
{\color{DarkRed}{All in all, with the rapid advance of this ﬁeld, \textbf{Generalization}, \textbf{Robustness}, and \textbf{Fairness} have been crucial aspects in federated learning. Generalization ensures federated performance on heterogeneous distribution. Robustness guarantees the federation to defend against malicious attacks. Fairness provides reasonable interest allocation to maintain sustainable multiple-party collaboration. Although there is a huge body of new literature, most existing surveys focus on the \textbf{narrow view} with \textbf{fragmented results}. In contrast, we argue that these three pieces interact with each other to jointly enhance the practical federation deployment and this is the first work to  \textbf{simultaneously investigate} the related research development and \textbf{uniformly benchmark} multi-view experimental analysis on the {Generalization}, {Robustness}, and {Fairness} realms.
}}

\subsection{Contribution}
\label{sec:contribution}
In this paper, we introduce a comprehensive survey on federated learning and mainly focus on generalization, robustness, and fairness. Compared with existing surveys, this paper makes the following contributions:
\begin{fullitemize}
\item We delve into an in-depth exploration of federated learning and provide the first state-of-the-art and symmetric survey on the generalization for heterogeneous distribution, robustness for malicious attacks, and fairness for multiple-party interest in federated learning, including hundreds of papers in this fast-growing ﬁeld.

\item We select inﬂuential works published in
prestigious journals and conferences and classify the existing federated methods, based on different task settings: \croclishift{} and \outclishift{} in \gfl{}, \bayatt{} and \bacatt{} in \rfl{}, \rewaconf{} and \predbias{} in \ffl{}. Except for the taxonomies, in-depth analysis about the pros and cons of these methods is also provided. 

\item  We conduct a plentiful benchmark analysis on various federated scenarios with different solutions. Alongside a set of evaluation metrics: generalizable performance, defensive degree, and fair level, we comprehensively investigate the methods effectiveness.

\item  We discuss future research directions, which will assist the community to rethink and improve their current designs toward federated learning in real practicality. Meanwhile, in order to promote the development of this field.

\end{fullitemize}
The remainder of this paper is organized as follows. \cref{fig:framework} shows the structure of this survey. \cref{sec:background} gives some brief introduction on the history, terminology, and problem definition. We review representative papers on federated algorithms from three aspects: \gfl{} in \cref{sec:gfl}, \rfl{} in \cref{sec:rfl}, and \ffl{} in \cref{sec:rfl}, respectively.  With respect to the experimental setting \cref{sec:setup}, we introduce the federated datasets, implementation details, and thoroughgoing evaluation metrics.
\cref{sec:benchmark} conducts performance evaluation and analysis. While {\cref{sec:future_direction}} raises open questions and potential directions. Finally, we make concluding remarks in \cref{sec:conclusion}.

\section{Background}
\label{sec:background}

\subsection{History and Terminology}
\label{sec:historyandteriminology}
Federated learning typically aims to enable systems or devices to collaboratively construct a global model while maintaining all private data locally, without the need for explicit data exchange among clients. This concept was introduced in milestone work such as \cite{FLOptimization_arXiv16,FLStrategies_arXiv16,FedAvg_AISTATS17}. Based on distribution characteristics, existing federated learning can be categorized into the following types {\cite{FLonNONIDD_NC21,FLThreat_IF23}}: 
{\hfl} (\hflabbrv): Clients own the same feature space while having different sample spaces, and collaboratively optimize a shared global model {\cite{FedAvg_AISTATS17,FedSeg_CVPR23}}.  
{\vfl} (\vflabbrv): Each device has a dataset with distinct feature characteristics derived from the same sample identification space \cite{VFL_arXiv22,VFLChallMethoExper_arXiv22}. Thus, \vfl{} can be seen as performing feature-level privacy-preserving computations.
{\ftl} (\ftlabbrv): When both sample spaces and feature spaces are different among clients, \ftlabbrv{} basically leverages the transfer learning technique to achieve knowledge communication \cite{FederatedTransferFrame_TS20,FTLConceptandApplication_IA21,FTLConceptandApplication_IA21}.
In this paper, we focus on three crucial characteristics in federated learning and we conduct the uniform benchmark evaluation on the  {\hfl{}} \footnote{We omit the word horizontal for brevity.} (\hflabbrv{}) scenario.
\textbf{I)} \textbf{\gfl} (\gflabbrv): As a slang goes that {when you hear hoofbeats, think of horses, not zebras}. During federated learning, the decentralized data poses non-independent and identically distribution (called data heterogeneity) \cite{ConvergenceofFedAvg_arXiv19,LinearSpeedupforParticalFL_ICLR21,FADA_ICLR20,FedDGSurvey_arXiv23}. Then, there naturally exist two types of shift. 
{\color{DarkRed}First, {\croclishift{}} denotes the difference between empirical risk loss among participating clients {\cite{FedLC_ICML22,FedProto_AAAI22,FedPCL_NeurIPS22}}.} 
Second, {\outclishift} means the difference in expected risk between participating and non-participating clients \cite{FADA_ICLR20,CCST_WACV23,FedINB_ICLR23}. 
\textbf{II)} \textbf{\rfl} (\rflabbrv): Recent works show that standard federated learning is vulnerable to malicious attacks. One stream, {\bayatt{}}, investigates polluting the local data or the uploaded model to jeopardize the discrimination and convergence of the federated model \cite{AdvML_SAI11,AsyByzantineML_ICML18,AFA_arXiv19}. The other group, {\bacatt{}} aims to induce the local model to incorporate misdefined information and activate the backdoor trigger to achieve malicious targeted prediction {\cite{BackdoorFL_NeurIPS19,HowtoBackDoorFL_AISTATS20,CanBackdoorFL_NeurIPS20,Ditto_ICML21}}.
\textbf{III)} \textbf{\ffl} (\fflabbrv): Fairness in federated learning mainly encompasses two aspects \cite{FedCE_CVPR23,FairandPrivacyFL_arXiv23,PrivacyrobustnessFL_TNNLS22}. During the training stage, clients need to optimize the distributed model on their local data. The local updating process incurs computation cost to train the model and the training effectiveness largely depends on the data quality \cite{FedCI_BigData19,PayoffandRewardCML_NeurIPS22}. Consequently, the computation cost and data value should be carefully considered to ensure the {\colfair}. As the saying goes \textit{No pain, no gain}, federated learning needs to estimate reward to incentivize the client to contribute to the federated system. Besides, given the distributed data distribution, fitting different client data poses varying difficulty degrees. Naturally, during the inference period, the shared model would manifest the biased performance of minority or marginalized groups \cite{AFL_ICML19,FCFL_NeurIPS21,FairFed_AAAI23}. Thus, maintaining {\perfair} becomes an essential objective to mitigate performance disparities. 

\subsection{Problem Definition}
\label{sec:prob}
Following the standard federated learning setup \cite{FedAvg_AISTATS17,FedProx_MLSys2020,MOON_CVPR21,FedLC_ICML22,FPL_CVPR23}, suppose there are $M$ clients, (indexed by $i$) with respective private data $D_i$. $N_i\!=\!|{D_i}|$ means the private data scale for the $i^{th}$ client private dataset. We further assume each data sample $(x,y)$, where $x$ is the input attribute and $y$ is the label. The local data follows the distribution $\mathbb{P}_i(x,y)$. We denote the global model parameter as $w \!= \! f \circ g$, where $f$ means the network backbone and $g$ represents the uniform classifier. For the query sample $\xi = (x,y)$, The $f$ maps $x$ into the $d$ dimensional feature vector $h = f(x) \in \mathbb{R}^d$. $g$ maps feature $h$ into logits output $z = g(h) \in \mathbb{R}^{|C|}$, where $C$ denotes classification set.
Formally, federated solutions generally seek to learn an \textit{ideal} global model, $w^*$ to minimize the  weighted empirical loss among clients:
\begin{equation}\small
\setlength\abovedisplayskip{2pt} \setlength\belowdisplayskip{2pt}
    w^* =  \min_w \sum_{i=1}^M\alpha_i F_i(w,D_i),
    \label{eq:originfl}
\end{equation}
where $\alpha_i$ denotes the pre-allocated aggregation weight ($\sum_{i=1}^N \alpha_i\!=\!1$) and is normally based on the data scale: $\alpha_i \! =\! \frac{N_i}{\sum_iN_i}$, or client scale: $\alpha_i \!=\! \frac{1}{M}$. $F_i(w,D_i)$ represents the empirical loss.
We define the $L_i$ as the client-specific loss function and further disassemble the federated process into the following three steps:
\begin{equation}\small
\setlength\abovedisplayskip{2pt} \setlength\belowdisplayskip{2pt}
\begin{aligned}
w_i \leftarrow{} w & &  \text{Distribute}\\
 \min_{w_i} \mathbb{E}_{\xi \in D_i}[L_i(w_i,\xi)]  & & \text{Optimize} \\
w = \sum_{i=1}^M  \alpha_i w_i & & \text{Aggregate}\\
\end{aligned}
\label{eq:flsteps}
\end{equation}


\subsubsection{Distributed Shift in FL}
\label{sec:distributedshift}
The isolated data distributions are often non-independent and non-identically distributed (non-IID), reflecting local distribution differences among clients {\cite{FedProx_MLSys2020,FedNova_NeurIPS20,FedAVGLinearSpeedUp_ICLR21,FedDG_CVPR21,SharpBoundsFL_AISTAS22}}. The local distribution $\mathbb{P}_i(x,y)$ can be considered as $\mathbb{P}_i(x|y)\mathbb{P}_i(y)$. 
According to how the distribution $\mathbb{P}_i(x,y)$ is defined, distributed shift can be broadly categorized into two classes: \croclishift{} and \outclishift{}.
\noindent$ \bullet$~\textbf{\croclishift{}}. In the distributed manner, different clients normally hold the local data with different distributions. Thus, there exists the cross-client shift, which brings the distinct local empirical risk minima. Thus, the divergent client updating direction hinders the federated convergence speed and weakens federated effectiveness. 

\noindent \textbf{i)}  \labelshift{}: The label distributions $\mathbb{P}_i(y)$ on the clients are distinct, and the conditional feature distribution $\mathbb{P}_i(x|y)$ is shared across clients. For instance, different hospitals specialize in distinct diseases and would hold inconsistent medical
records ratio. Experiments normally leverage the Dirchlet sampling {\cite{Dirichlet_2004}} to mimic the \labelshift{}. Furthermore, the \labelshift{} is formulated as:
\begin{equation}\small
\setlength\abovedisplayskip{2pt} \setlength\belowdisplayskip{2pt}
    \begin{split}
    \mathbb{P}_i(y) \! &\neq \! \mathbb{P}_j(y)\\
    \mathbb{P}_i(x|y) \! & =\! \mathbb{P}_j(x|y),
    \end{split}
    \label{eq:labelshift}
\end{equation}
where $i$ and $j$ denotes arbitrary clients in federation.

\noindent \textbf{ii)} \domainshift{}: As for the same label space, distinct feature distribution exists among different participants \cite{FedBN_ICLR21,FPL_CVPR23,FSMAFL_ACMMM22}. For example, cats may vary in coat colors and patterns in different body parts and we provide the mathematical definition in the following form:
\begin{equation}\small
\setlength\abovedisplayskip{2pt} \setlength\belowdisplayskip{2pt}
    \begin{split}
    \mathbb{P}_i(x|y)\! &\neq \! \mathbb{P}_j(x|y)\\
    \mathbb{P}_i(y) \! & =\! \mathbb{P}_j(y).
    \end{split}
    \label{eq:domainshift}
\end{equation}
\noindent$\bullet$~\textbf{\outclishift{}}. As shown in \cref{eq:originfl}, the federated system fits and promotes performance in cross-client distribution. But when deploying the federated model to outer client distribution (\ie, unseen domains), denoted as $o$, nonparticipating client data exhibits the distributional shift compared to the participating client private data. Therefore, the performance drop is commonly observed \cite{GeneralizationinFL_ICLR22} because it fails to consider the underlying domain shift on the unseen distribution, an issue named \outclishift{}. In our work, we conduct experiments based on the feature-level distribution skew and provide the following definition:
\begin{equation}\small
\setlength\abovedisplayskip{2pt} \setlength\belowdisplayskip{2pt}
\begin{split}
\mathbb{P}_i(x|y)\! &\neq \! \mathbb{P}_o(x|y)\\
\mathbb{P}_i(y) \! & =\! \mathbb{P}_o(y).
\end{split}
\label{eq:outclientshift}
\end{equation}
\subsubsection{Malicious Behavior in Federated Learning}
\label{sec:maliciousbehavior}
Participants are uncontrollable in the distributed setting, some of them are malicious and aim to destroy the federated paradigm. We broadly divide adversarial behavior into two types, based on their destroying goal. As for \bayatt{} {\cite{Krum_NeurIPS17,Bulyan_ICML18,AsyByzantineML_ICML18}}, it is untargeted attack and the adversary objective is to hinder the model from achieving a near-optimal performance on the major task. As for \bacatt{} {\cite{TargetBackdoor_arXiv17,BackdoorviaInvisiblePerturbation_arXiv18,BadNets_Access19}}, it could be assumed as the targeted attack, which ensures the optimized model presents differently on certain targeted sub-tasks while maintaining good overall performance on the primary task.

\noindent$\bullet$~\textbf{\bayatt{}{}}.
Baleful client uploads the modified information to inhibit federated convergence and performance from the data-based and parameter-based angles.

\noindent \textbf{i)} \databayatt{}.
As for the data-based attacks {\cite{AdvML_SAI11,PoisoningattacksSVM_ICML12}}, malicious clients would intentionally pollute the local data to corrupt the learned model. We mainly consider the following two kinds and $\epsilon$ denotes the noise rate that the label is flipped from the clean to the noisy class:
\begin{fullitemize}
\item\symflip{} (\symflipabbrv{}) {\cite{SymmetryFlip_NeurIPS15}}: The original label will be flipped to any wrong classes with the equal ratio. \begin{equation}\small
\setlength\abovedisplayskip{2pt} \setlength\belowdisplayskip{2pt}
\text{\symflipabbrv{}} = \left[ \begin{array}{cccc}
1-\epsilon & \frac{\epsilon}{|C|-1} & \cdots& \frac{\epsilon}{|C|-1} \\
\frac{\epsilon}{|C|-1} & 1-\epsilon & \cdots  & \frac{\epsilon}{|C|-1} \\
\vdots & \vdots & \ddots & \frac{\epsilon}{|C|-1} \\
\frac{\epsilon}{|C|-1} & \frac{\epsilon}{|C|-1} & \frac{\epsilon}{|C|-1}  & 1-\epsilon
\end{array} 
\right].
\label{eq:symflip}
\end{equation}
\item \pairflip{} (\pairflipabbrv{}) {\cite{PairFlip_NeurIPS18}}: The original class label would only be flipped to a  similar wrong semantic.
\begin{equation}\small
\setlength\abovedisplayskip{2pt} \setlength\belowdisplayskip{2pt}
\text{\pairflipabbrv{}} = \left[ \begin{array}{cccc}
1-\epsilon & \epsilon & \cdots& 0 \\
0 & 1-\epsilon & \cdots  & 0 \\
\vdots & \vdots & \ddots & \epsilon \\
\epsilon & 0 & 0 & 1-\epsilon
\end{array} 
\right ].
\label{eq:pairflip}
\end{equation}
\end{fullitemize}
\noindent \textbf{ii)} \modelbayatt{}.
With respect to parameter-based {\cite{AdvML_SAI11,BackdoorFL_NeurIPS19,AdvLens_ICML19,DBA_ICLR20,HowtoBackDoorFL_AISTATS20,IPM_UAI20,CanBackdoorFL_NeurIPS20,IPM_UAI20}}, evil participants manipulate uploading model parameters before sharing with the server. We utilize the uploading gradient of the $k$ participant as an example. For kind clients, they faithfully upload the gradient change as $\nabla_k$. But malicious clients deliberately send distorted signals. We majorly consider the following attacks types:
\begin{fullitemize}
\item \randomnoise{} (\randomnoiseabbrv{}{}): Straightforwardly modify the neural network via random sampling values as:
\begin{equation}\small
\setlength\abovedisplayskip{2pt} \setlength\belowdisplayskip{2pt}
\nabla_k  = *.
\label{eq:randomnoise}
\end{equation}
$*$ denotes the arbitrary values and normally leverages Gaussian Distribution or default initialization function to generate the parameter distortion.
\item \lie{} (\lieabbrv{}) \cite{LIE_NeurIPS19}: Assume the complete knowledge of the gradients of benign clients. Add a very limited amount of noise to aggregation. This kind of distortion is significant enough to negatively affect the global model but subtle enough to evade detection by the \byzantoler{} solutions. Specifically, it calculates the average value $\mu$ and standard deviation $\sigma$ on the benign ones. Then, measure the coefficient $\textit{z}$ on the overall clients. The malicious updates is formed as $\mu +\textit{z} \sigma$.

\item \fang{} \cite{Fang_USENIX20}:  Formulate the attacks as optimization-based model poisoning tailored to different defensive methods. Specifically, \fang{} crafts the local models on the compromised clients such that
the global model deviates the most towards the inverse of the
the direction along which the previous normal global model would
change.
\item \minmax{} (\minmaxabbrv{}) \cite{DnC_NDSS21}: Ensure that the evil gradients lie close to the benign gradient group. We calibrate the malicious gradient to ensure that its maximum distance from any other gradient is limited by the maximum distance between benign gradients as the following form:
\begin{equation}\small
\setlength\abovedisplayskip{2pt} \setlength\belowdisplayskip{2pt}
\begin{aligned}
\arg \max_\gamma  & \max_{i \in [n]} ||\nabla_k -\nabla_i||_2 \leq \max_{i,j \in [n]} ||\nabla_i -\nabla_j||_2,\\
& \nabla_k = AVG(\nabla_{\{i \in [n]\}})+\gamma\nabla^p,
\end{aligned}
\label{eq:minmaxattack}
\end{equation}
where $\nabla^p$ means the perturbation vector, $\gamma$ is the learnable scaling coefficient and $[n]$ is the benign client clique. 
\item \minsum{} (\minsumabbrv{}) \cite{DnC_NDSS21}:
The objective is to ensure that the sum of squared distances between the malicious gradient and all benign gradients remains below an upper bound, smaller than the sum of squared distances between any benign gradient and the other benign gradients as:
\begin{equation}\small
\setlength\abovedisplayskip{2pt} \setlength\belowdisplayskip{2pt}
\begin{aligned}
\arg \max_\gamma  & \sum_{i \in [n]} ||\nabla_k -\nabla_i||_2 \leq \max_{i\in [n]} \sum_{j \in [n]} ||\nabla_i -\nabla_j||_2,\\
& \nabla_k = AVG(\nabla_{\{i \in [n]\}})+\gamma\nabla_p.
\end{aligned}
\label{eq:minsumattack}
\end{equation}

\end{fullitemize}
\noindent$\bullet$~\textbf{\bacatt{}}.  Compared with direct performance degradation, \bacatt{} focuses on injecting a backdoor trigger pattern into the existing model while retaining the major task accuracy \cite{BackdoorFL_NeurIPS19,AnalyzeFLthroughAdvLens_ICML19,CBD_CVPR23}. Specifically, we define the trigger pattern as the $\bm{\Phi}$ and the modified sample as $\widetilde{x}=(1-\textbf{m}) \odot x + \textbf{m} \odot \bm{\Phi}$. Thus, the original local optimization direction in \cref{eq:flsteps} would be reformulated into:
\begin{equation}\small
\setlength\abovedisplayskip{2pt} \setlength\belowdisplayskip{2pt}
\min_{w_i} \mathbb{E}_{(x,\widetilde{x},y) \in D_i}[L_i(w_i,x,y)+  {\varpi \underbrace{L(w_i,\widetilde{x},\widetilde{y})}_{Backdoor}}].
\label{eq:backdoor}
\end{equation}
The $\widetilde{y}$ means the preset attack target predictions.
The $\textbf{m}$ means the trigger location mask.
The hyper-parameter $\varpi$ is utilized to control the trade-off between the original common empirical risk minimization and the backdoor injection loss function. Prior backdoor attacks are normally motivated by centralized backdoor attacks and assume that each malicious attacker independently trains their local models, without any collusion among them {\cite{HowtoBackDoorFL_AISTATS20,CanBackdoorFL_NeurIPS20}. Therefore, they utilize the same backdoor trigger, the poisoned local models tend to appear similar updating trend, largely deviated from the benign optimizing direction \cite{FoolsGold_arXiv18}. Recently, advanced attacks focus on the distributed backdoor paradigm \cite{DBA_ICLR20,CerP_AAAI23,F3BA_AAAI23,F3BA_AAAI23} and dynamic backdoor solution {\cite{DynamicBackdoorFL_arXiv20,LearntoBackdoorFL_ICLRW23}}} to evade the secure detection strategies. 

\subsubsection{Interest Conflict in Federated Learning}
\label{sec:interestconflict}
Unlike the traditional centralized training paradigm, which optimizes on local data, federated learning involves multiple clients collaboration. Thus, as a saying goes \textit{No rules, no justice}. The federation needs to carefully deal with the client interest conflict to satisfy the expected performance improvement and reasonable benefit allocation \cite{ExecutionAssuranceofFL_INFOCOM20,FAFL_TNNLS23}. We mainly discuss several types of interest conflict:

\noindent$\bullet$~\textbf{\rewaconf{}{}}.
Existing federated literature is normally under the assumption that data owners have an inherent willingness to participate in the federated system \cite{FedAvg_AISTATS17,FedProx_MLSys2020,LGFEDAVG_NeurIPS20}. However, during the federated process, the contributions of various clients are not uniform due to differences in local data value and computational costs \cite{SDRCGM_AAAI22}. Consequently, it becomes necessary to ensure that the client reward is proportionate to its contribution to the federation. Specifically, a participant contributing significantly should receive a higher reward than one with a lower contribution. Normally, existing research leverages the Game theory \cite{GameTheory_97}, which investigates the strategic interactions among rational agents. In our work, we utilize the Shapley Value \cite{Shapley_97}, which is a marginal contribution-based scheme from cooperative game theory \cite{CooperativeGame_12,KernelofCooperativeGame_65} to evaluate the client contribution. To be precise, we consider the standard test dataset $u$ for performance evaluation. We denote the online client set $S \subseteq M$ and  $D_S$ as the union dataset set. The overall accuracy on testing set $u$ accuracy of federated model $w$, optimized on $D_S$, is defined as $A(w, D_S,u)$, abbreviated as $A_S^u$. The  Shapley value for the client $i$, $\nu_i(w,D_S,D_i,u)$, abbreviated as $\nu_i$, is formulated as the following form:
\begin{equation}\small
\setlength\abovedisplayskip{2pt} \setlength\belowdisplayskip{2pt}
\nu_i = \frac{\rho}{|M|}  {\sum_{S \subseteq M \setminus \{ i\}} \frac{A_{S\cup \{i\}}^u - A_S^u}{\bigl(\begin{smallmatrix} |M|-1 \\ |S|  \end{smallmatrix}\bigr)} },
\label{eq:shapleyvalue}
\end{equation}
where $\rho$ means the constant parameter to rescale the shapely value. Via the \cref{eq:shapleyvalue}, we assess the importance of client contribution via evaluating its marginal effect for the respective ﬁnal model accuracy.

\noindent$\bullet$~\textbf{\predbias{}{}}. Due to the data heterogeneity in the federated paradigm, it is possible that the federated performance would vary significantly with uneven performance on different distributions \cite{AFL_ICML19,qFFL_ICLR20}. For instance, under the \domainshift{} (\cref{eq:domainshift}), private data is normally derived from different domains with diverse distribution, \eg, \mnist{} \cite{MNIST_IEEE98} and \svhn{} \cite{svhn_NeurIPS11} in the \digits{} scenario. Regrettably, solving the optimization in \cref{eq:originfl} does not inherently promote the shared global model to produce uniform predictive accuracy across multiple domains, referred to as \predbias{}. Specifically, due to the fitting difficulty and similarity correlation \cite{FEDCG_CVPRW21}, the federated model would be biased towards specific domains and thus present inferior performance on those domains with hard or distinct distribution with others \cite{FedFV_IJCAI21,FairFed_AAAI23}. We mathematically define the metric to evalutate the \predbias{} degree as:  
\begin{equation}\small
\setlength\abovedisplayskip{2pt} \setlength\belowdisplayskip{2pt}
\zeta = \sigma \{{\bm{\mathcal{A}}}^u\}_{u \in {\mathcal{U}}}
\label{eq:prebias}
\end{equation}
The $\sigma$ means the standard deviation and ${\mathcal{U}}$ denotes the testing dataset collection.
We assume that all clients join the federation and omit the $A_M^T$ as $A^T$. The larger the $\zeta$ is, the more biased the predictive results are. 
\section{\gfl}
\label{sec:gfl}
\subsection{Generalization Metrics}
\label{sec:gen_eval_metrics}
{\color{DarkRed}
To achieve the generalization property during federation, existing methodologies for \croclishift{} and \outclishift{} problems, are divided into two major streams: \crosscal{} (\cref{sec:crosscal}) and \ungene{} (\cref{sec:ungene}). 
\noindent$\bullet$~\textbf{\croclienacc{}: ${\bm{\mathcal{A}}}^{\mathcal{U}}$}. As for the \croclishift{} methods \cref{sec:crosscal}, it mainly considers that the client distribution is sampled from the single domain distribution, \labelshift{} or multiple domain distributions, \domainshift{}. Thus, when it comes to the evaluation stage, we abstract it as a test dataset collection with an uncertain scale, ${\mathcal{U}}$. We default assume that clients are sampled from several source domains and there is a testing dataset collection ${\mathcal{U}}\!=\!\{u\}$ and the unknown testing dataset $O$. We acquire the logits output ($z \!=\! w(x)$) from the global model on the query testing sample, $\xi\!=\!(x,y)$. We measure the standard Top-1 accuracy metric on each specific testing dataset $u \in {\mathcal{U}}$. We further utilize the average value to represent the performance in the following formulation:
\begin{equation}\small 
\setlength\abovedisplayskip{2pt} \setlength\belowdisplayskip{2pt}
\begin{aligned}
{\bm{\mathcal{A}}}^{u} &= \frac{\sum ( \max(z)=y)}{|u|},\\
{\bm{\mathcal{A}}}^{\mathcal{U}} & = \frac{1}{|{\mathcal{U}}|} \sum_{u \in {\mathcal{U}}}
{\bm{\mathcal{A}}}^{u}.
\end{aligned}
\label{eq:cross_acc}
\end{equation}

\noindent$\bullet$~\textbf{\outclienacc{}: ${\bm{\mathcal{A}}}^{O}$}. Similarly, with respect to the \outclishift{} solutions \cref{sec:ungene}, it focuses on evaluating the unknown distribution. Thus, we directly measure the Top-1 accuracy metric on the unseen domain distribution $O$ as the following formulation:
\begin{equation}\small 
\setlength\abovedisplayskip{2pt} \setlength\belowdisplayskip{2pt}
{\bm{\mathcal{A}}}^{O} = \frac{\sum (\max (z)=y)}{|O|}.
\label{eq:out_acc}
\end{equation}
}
\subsection{\crosscal{}} 
\label{sec:crosscal}
As for the \croclishift{} problem, each local distribution is highly heterogeneous with others, and the local objective of each party is inconsistent among clients. Thus, different client optimizes towards distinct local empirical minimization, bringing the divergent optimization direction Related methods focus on calibrating the client optimization direction via three major angles.

\subsubsection{\clientreg{}}
\label{sec:clientreg}

\noindent$\bullet$~\textbf{Global Neural Network.} The most intuitive method is to
directly use a shared global model for local optimization guidance because it literally aggregates clients knowledge and naturally acts as the global knowledge representative. Some works focus on directly regulating the local network based on the parameter stiff \cite{FedProx_MLSys2020,FedCurv_NeurIPSW19,FedDyn_ICLR21} and parameter variance \cite{SCAFFOLD_ICML20,FedGA_AAAI22,FedDC_CVPR22,FedPVR_CVPR23,GradMA_CVPR23}. A handful of methods utilize the global model output on the private data to penalize the local direction 
{\color{DarkRed}\cite{FLwiMoE_2020,MOON_CVPR21,FedUFO_ICCV21,FedNTD_NeurIPS22,FedSeg_CVPR23,FedNoRo_IJCAI23}. However, involving a global model in the local optimization process deeply enlarges the local computation cost and linearly increases with the parameter scale. }

\begin{table}[t]\small
\caption{
\small{\textbf{Summary of essential characteristics for reviewed methods in \crosscal{}} (\cref{sec:crosscal}).}
}
\label{tab:computation}
\vspace{-10pt}
\centering
{
\resizebox{\columnwidth}{!}{
\setlength\tabcolsep{1pt}
\renewcommand\arraystretch{1.2}
\begin{tabular}{rIc|c}
\hline\thickhline
\rowcolor{mygray}
Methods & Venue  & Highlight \\
\hline\hline 
\multicolumn{3}{l}{\textit{Global Neural Network}}\\  
\multicolumn{3}{l}{\textbf{Limitation}: Linearly increase the local computation cost}\\ 
\hline
\fedprox{}\cite{FedProx_MLSys2020} & \pub{MLSys'20} & Parameter $\ell_2$ normalization\\
\scaffold{}\cite{SCAFFOLD_ICML20} & \pub{ICML'20} &  Control variate\\
\moon{}\cite{MOON_CVPR21} & \pub{CVPR'21} &  Feature-level contrastive learning \\
\fedntd{}\cite{FedNTD_NeurIPS22} & \pub{NeurIPS'22} & Decoupled distillation \\
\fedseg{}\cite{FedSeg_CVPR23} & \pub{CVPR'23}  & Pixel-level contrastive learning\\
\hline\hline 
\multicolumn{3}{l}{\textit{Global Statistic Information}} \\  
\multicolumn{3}{l}{\textbf{Limitation}: Reliance on data richness for ideal initialization}\\ 
\hline
\fedproc{}\cite{FedProc_arXiv21} & \pub{arXiv'21} & Prototypical contrast\\
\harmofl{} \cite{HarmoFL_AAAI22} & \pub{AAAI'22} & Normalized  amplitude\\
\fedfa{} \cite{FedFA_ICLR23} & \pub{ICLR'23} & Gaussian augmentation \\
\fpl{}\cite{FPL_CVPR23} & \pub{CVPR'23} & Cluseter and unbiased prototypes\\
\hline\hline 
\multicolumn{3}{l}{\textit{Extra Network Architecture}} \\ 
\multicolumn{3}{l}{\textbf{Limitation}: Compatibility difficulty and communication burden}\\ 
\hline
\fedmlb{}\cite{FedMLB_ICML22} & \pub{ICML'22}  & Multiple auxiliary branches\\
\fedcgan{}\cite{FedCG_IJCAI22}  & \pub{IJCAI'22}  & Conditional GAN\\
\adcol{}\cite{AdCol_ICML23}  & \pub{ICML'23} & Representation generator\\
\dafkd{}\cite{DaFKD_CVPR23} & \pub{CVPR'23}  & Discriminator module\\
\hline\hline 
\multicolumn{3}{l}{\textit{Self-Driven Regularization}} \\  
\multicolumn{3}{l}{\textbf{Limitation}: Forgetting with unstable hyper-parameter}\\ 
\hline
\fedrs{}\cite{FedRS_KDD21} & \pub{KDD'21} & Restrict Softmax function\\
\fedalign{} \cite{FedAlign_CVPR22} & \pub{CVPR'22} & Final block Lipschitz constant  \\
\fedsam{} \cite{FedSAM_ICML22} & \pub{ICML'22} & Sharpness aware minimization\\
\fedlc{} \cite{FedLC_ICML22} & \pub{ICML'22} & Calibrate logit via class probability \\
\feddecorr{} \cite{FedDecorr_ICLR23} & \pub{ICLR'23} & Dimensional decorrelation \\
\hline\hline 
\multicolumn{3}{l}{\textit{Federated Data Sharing}} \\  
\multicolumn{3}{l}{\textbf{Limitation}: Pre-defined relevant data is not always available}\\ 
\hline
\dcadam{} \cite{DCAdam_CS21} & \pub{CS'21} & Pre-shared data for warm-up strategy\\
\fedaux{} \cite{FEDAUX_TNNLS21} & \pub{TNNLS'21} & Pretrain and distill  on auxiliary data \\
\hline\hline 
\multicolumn{3}{l}{\textit{Federated Data Enhancement}} \\  
\multicolumn{3}{l}{\textbf{Limitation}: Limited diversity and potential privacy leakage}\\ 
\hline
\fedmix{}\cite{FedMix_ICLR21} & \pub{ICLR'21} &  Close global Mixup via averaged data\\
\fedgen{}\cite{FEDGEN_ICML21} & \pub{ICML'21} & Data generator ensemble learning\\
\hline\hline 
\multicolumn{3}{l}{\textit{Federated Data Selection}} \\  
\multicolumn{3}{l}{\textbf{Limitation}: Data and client-level unfairness}\\ 
\hline
\fedacs{}\cite{FedACS_IWQOS21} & \pub{IWQOS'21} & Filter out poisoned data via clustering\\
\safe{}\cite{Safe_TII22} & \pub{TII'22} & Thompson sampling low skew clients \\
\hline\hline 
\multicolumn{3}{l}{\textit{Server Aggregation Reweighting}} \\  
\multicolumn{3}{l}{\textbf{Limitation}: Require qualified datasets and element evaluation}\\ 
\hline
\fedbe{}\cite{FEDBE_ICLR21} & \pub{ICLR'21}  & Bayesian ensemble inference \\
Elastic \cite{ElasticAgg_CVPR23} & \pub{CVPR'23} & Interpolate parameter via sensitivity\\
\hline\hline 
\multicolumn{3}{l}{\textit{Server Adaptive Optimization}} \\  
\multicolumn{3}{l}{\textbf{Limitation}: Demand proxy dataset and elaborate joint-goal}\\ 
\hline
\fedmd{} \cite{FedMD_NeurIPS19} & \pub{NeurIPS'19} & Additional classifier for  data distillation\\
\feddf{} \cite{FedDF_NeurIPS20} & \pub{NeurIPS'20} & Ensemble distillation on related data\\
\fedgkt{}\cite{FedGKT_NeurIPS20} & \pub{NeurIPS'20}  & Server-side group knowledge transfer \\
\fedopt{}\cite{FedOPT_ICLR21} & \pub{ICLR'21} & Federated  adaptive server optimizer\\
\fccl{} \cite{FCCL_CVPR22} &  \pub{CVPR'22} &  Cross-Correlation on unrelated data\\
\end{tabular}}}
\vspace{-10pt}
\end{table}

\noindent$\bullet$~\textbf{Global Statistic Information.} Although the global model represents the overall federated knowledge, it fails to provide more fine-grained information.  A group of federated methods has developed to construct the class-wise representative, \eg prototype  \cite{FedProc_arXiv21,FPL_CVPR23} and Gaussian distribution \cite{CCVR_NeurIPS21,FedFA_ICLR23}. Some approaches make use of spectrum \cite{HarmoFL_AAAI22} and feature maps \cite{FedTFI_SCIS22}.  
However, it is vital to note that generating reliable global statistical knowledge necessitates data enrichment. Without this preliminary, initial signals may introduce bias and result in biased federated models.

\noindent$\bullet$~\textbf{Extra Network Architecture.} A plethora of works switch to utilizing additional network structures such as generative adversarial networks \cite{GAN_NeurIPS14,FEDGEN_ICML21,FedCG_IJCAI22,DaFKD_CVPR23} and auxiliary global branch \cite{FedGKT_NeurIPS20,FedMLB_ICML22,FLwiMoE_2020}. Nevertheless, they largely restrict the participating network architecture selection scope, which needs to be compatible with the extra network architecture and increase the communication cost.

\noindent$\bullet$~\textbf{Self-Driven Regularization.}  A stream calibrates the client drift by enhancing the local learning generality in a self-driven manner. Generally, it adopts the self-distillation paradigm \cite{Fed2_SIGKDD21,FedAlign_CVPR22} or modified Cross Entropy term \cite{CE_AOR05,FedRS_KDD21,FedLC_ICML22,FedDecorr_ICLR23}. Although these methods get rid of the additional shared signals cost, they are normally hyper-parameter sensitive and presents fragile under serious data heterogeneous scenarios.

\subsubsection{\clientaug{}}
\label{sec:clientaug}

\noindent$\bullet$~\textbf{Federated Data Sharing.} To alleviate the damage of heterogeneous client data, directly sharing relevant datasets acts as a naive and simple solution during federation. Current solutions normally leverage the labeled samples for the warm-up operation \cite{FLwithNonIID_arXiv18,DCAdam_CS21} and unlabeled instances for distillation {\cite{DSFL_TMC21}} and pretraining {\cite{FEDAUX_TNNLS21}} strategies. However, in scenarios characterized by data scarcity, it may be infeasible to select suitable related datasets for sharing.

\noindent$\bullet$~\textbf{Federated Data Enhancement.} The client shift problem originates from the inconsistency between the ideal global and local client distribution. A surge of efforts supplements local data to mimic the ideal data distribution via  the batch normalization layer \cite{FedZDAC_CVPRW21,BatchNorm_ICML15}, MixUp {\cite{FedMix_ICLR21,Mixup_ICLR18,MixUPBenefitOverConfidence_NeurIPS19,Astraea_ICCD19,XorMixup_arXiv20}}, data generative generator {\cite{FAug_arXiv18,FedZDAC_CVPRW21}}. However, federated data enhancement largely depends on the local data diversity and exists the privacy leakage potential.

\noindent$\bullet$~\textbf{Federated Data Selection.} Another direction proposes dynamic data selection without impacting the data collection and client optimization. Generally speaking, this approach either employs local data filtering strategies \cite{Safe_TII22,FEDSS_ICC20} to select meaningful data while rejecting potentially tainted samples, or {\color{DarkRed} it establishes client selection paradigms \cite{MinCost_TPDS20,FAVOR_INFOCOM20,FedACS_IWQOS21,FedAgg_IWQOS21,FedCor_CVPR22,PowerofChoice_AISTATS22} that prioritize those with lower degrees of data heterogeneity. However, it is worth noting that while federated data selection can significantly boost convergence speed, it unavoidably introduces unfairness at both the data and client levels since it tends to favor the selection of mainstream data or clients while potentially ignoring other diversified samples.}

\subsubsection{\serverope}
\label{sec:serverope}

\noindent$\bullet$~\textbf{Server Aggregation Reweighting.} Compared to modifying local training paradigms for calibrating client shift, existing research has confirmed that default fixed parameter aggregation ($\alpha_i$ in \cref{eq:originfl}) also leads to slow convergence and accuracy drop because it fails to effectively calibrate client divergence \cite{NonIIDQuagmireofFL_ICML20}. Thus, a dynamic parameter aggregation framework is helpful to improve federated convergence and performance. Related solutions can be mainly divided into several types. Some techniques utilize different post-processing methods to mitigate the model drift, \eg, Bayesian non-parametric theory \cite{BNPFL_ICML19,SPAHM_NeurIPS19,FedMA_ICLR20,FEDBE_ICLR21}, MCMC \cite{FEDPA_ICLR21}, optimal transport \cite{OT_NeurIPS20}, gradient-based optimization \cite{AutoFedAvg_arXiv21,SmartFL_arXiv22}, and shapley-value \cite{ShapleyFL_SIGKDD23}. Another big family is built upon network characteristics \cite{DivFL_ICLR21,FedSim_NC22,ElasticAgg_CVPR23,FEDLAW_ICML23}. However, they either 
require unlabeled auxiliary datasets in the server or involve element-wise parameter evaluation, which is time-consuming and laborious. 

\noindent$\bullet$~\textbf{Server Adaptive Optimization.} 
Notably,  adaptive optimization has proven to be an effective solution for tailoring models to specific tasks. One approach focuses on server-side fine-tuning by synthesizing relevant information to facilitate empirical risk minimization \cite{SynDatasets_arXiv19,FedFTG_CVPR22,CReFF_ICJAI22,FedDM_CVPR23,DYNAFED_CVPR23,FedGKT_NeurIPS20} or conducting the knowledge distillation on public dataset for knowledge transferring \cite{ModelComp_KDD06,LUPI_JMLR15,KD_arXiv15,UnifDis_ICLR16,FedMD_NeurIPS19,FedDF_NeurIPS20,FedMDNFDP_IJCAI21,FCCL_CVPR22}. Another stream investigates the server-side optimizing objective, including gradient-biased server optimizer \cite{FedOPT_ICLR21} and global local joint objectives \cite{IDA_MICCAIW20,FedSAE_IJCNN21,FedAda_WWW22}. Clearly, such implicit target-appearance modeling strategy hurts ﬂexibility and adaptivity due to the requirement for the proxy dataset or elaborate optimization definition.

\begin{table}[t]\small
\caption{\textbf{Summary of essential characteristics for reviewed solutions for \ungene{} (\cref{sec:ungene})}. $\star$ and $\circ$ respectively represent the \textbf{potential privacy leakage} and \textbf{network architecture modification}.}
\label{tab:summaryUnknown Generalization}
\vspace{-10pt}
\centering
{
\resizebox{\columnwidth}{!}{
\setlength\tabcolsep{1pt}
\renewcommand\arraystretch{1.2}
\begin{tabular}{rIc|c|l}
\hline\thickhline
\rowcolor{mygray}
Methods & Venue & Highlight & Limitation\\
\hline\hline
\multicolumn{4}{l}{\textit{Federated Domain Adaptation \cref{sec:fda}}} \\  
\hline
\fada{} \cite{FADA_ICLR20}  & \pub{ICLR'20} & Dynamic adversarial alignment  & $\circ$: GAN \cite{GAN_NeurIPS14} \\ 
\copa{} \cite{COPA_ICCV21}  & \pub{ICCV'21} & Shared extractor \& group heads & $\circ$: IBN \cite{IBN_ECCV18} \\ 
\aegr{} \cite{AEGR_ICME23}  & \pub{ICME'23} & Pseudo-label refinement & $\star$: PGD \cite{PGD_arXiv17} \\ 
\hline\hline
\multicolumn{4}{l}{\textit{\fdg{} \cref{sec:fdg}}} \\  
\hline
\feddg{} \cite{FedDG_CVPR21}  & \pub{CVPR'21} & Sharing amplitude spectrum & $\star$: Amplitude \\ 
\ccst{} \cite{CCST_WACV23} & \pub{WACV'23} & Cross-client style transfer & $\star$: Style info \\
\csac{} \cite{CSAC_TKDE23} & \pub{TKDE'23} & Layer semantic aggregation & $\circ$: Attention
\end{tabular}}}
\vspace{-10pt}
\end{table}

\subsection{\ungene{}}
\label{sec:ungene}
Previous research has demonstrated that deep neural networks tend to empirically overfit training data, leading to over-confident predictions \cite{CalibrationDNN_ICML17,UncertainEstimateDeepEnsembles_NeurIPS17}. This phenomenon can be problematic in real-world scenarios \cite{ConcreteProblem_arXiv16}, where even minor variations,\ie domain shift, in the characteristics of deployment examples can result in significant performance drop \cite{PrediUncertainty_NeurIPS17,PrediUncertainty_Datashift_NeurIPS17}. With respect to the federated system, most existing works
only focus on improving model performance among clients, while ignoring model generalizability onto unseen domains outside the federation \cite{FLwithDA_NeurIPS19,FADA_ICLR20,FedDG_CVPR21,GeneralizationinFL_ICLR22}. Existing efforts can be mainly divided into two groups based on the accessible stage of the unseen domain: \fda{} (\fdaabbrv{} \cref{sec:fda}) and \fdg{} (\fdgabbrv{} \cref{sec:fdg}).

\subsubsection{\fda{}}
\label{sec:fda}
There has been an explosive interest in the \fda{} (\fdaabbrv{}) paradigm. \fdaabbrv{} normally assumes to incorporate unlabeled data from target domains during the federated learning process. Closely related methods mainly divide into two factions.

\noindent$\bullet$~\textbf{Federated Domain Alignment.}  The naive solution for handling domain shift with the target domain is to directly align and harmonize the multiple domain distributions. Relevant methods typically incorporate the contrastive learning \cite{MCCDA_TCSVT22,IRCMSDA_ICASSP23}, knowledge distillation {\cite{KD3A_ICML21,MCKD_ICASSP23,CoMDA_TCSVT23}}, adversarial learning {\cite{AEGR_ICME23}},  gradient matching {\cite{FedILC_arXiv22,GMFedDA_TNNLS22}}. In essence, this stream helps to align the representations of different domains, reducing the domain shift and thus enhancing the model generalizable ability across domains. However, a significant challenge stems from the problem setup, wherein the target domain is used during optimization but may not be accessible in many real-world scenarios.

\noindent$\bullet$~\textbf{Federated Domain Disentanglement.}  This line of methods focuses on separating knowledge into domain-invariant and domain-specific components. Speciﬁcally, \cite{FADA_ICLR20,AdvML_SAI11} explore the extension of adversarial adaptation techniques to the federated setting.  \cite{COPA_ICCV21} explicitly disentangle domain-invariant feature extractor and the ensemble multiple domain-speciﬁc classiﬁers. \cite{FLwiMoE_2020} utilize the Mixture of Experts theory \cite{EvaluateMoE_NeurIPS90,MoE_AIR14} to adjust the domain expert and domain generalist via gating mechanism. Nevertheless, it is important to note that these methods often rely on additional networks or the preservation of specialized modules to achieve the decoupling objective.



\subsubsection{\fdg{}}
\label{sec:fdg}
\fdg{} investigates deploying the models trained on sites of heterogeneous distributions, directly generalizing to unknown target clients with domain shift. Several pioneering works attempt to incorporate domain generalization problems into federated learning and can be generally categorized into two classes:

\noindent$\bullet$~\textbf{Federated Invariant Optimization.} 
A major stream of methods exploits calibrating the local training objective to alleviate the domain shift influence. Closely related solutions normally leverage the shared information, \eg, amplitude spectrum  \cite{FedDG_CVPR21}, style distribution \cite{CCST_WACV23} and Iterative Naive Barycenter \cite{FedINB_ICLR23} to learn domain invariant representation. Obviously, sharing data could be regarded as a breach of privacy \cite{FedSR_NeurIPS22,CCST_WACV23} and leads to a significant communication burden. Besides, a cluster of methods harness the additional modules or replace original network architecture to enhance the network generality and alleviate the domain biased trend, such as GAN \cite{GAN_NeurIPS14} in \fedadg{} \cite{FedADG_arXiv21,FedADG_IOTJ23}, Adain \cite{AdaIN_CVPR17} in \ccst{} \cite{CCST_WACV23}, and IBN \cite{IBN_ECCV18} in \copa{} \cite{COPA_ICCV21}.  However, implementing these approaches can be complicated because of adapting to various structures.

\noindent$\bullet$~\textbf{Federated Invariant Aggregation.} Yet another complementary line of work tries to modify the aggregation strategy to enhance the domain invariance ability. Some take inspiration from performance fairness to achieve multip-domain performance fairness in order to mitigate the generalization gap via the empirical variance balance \cite{FedGA_CVPR23}, agnostic
distribution fusion \cite{FedFusion_CVPR23}. A crowd of works delves into parameter aggregation from a fine-grained perspective, such as layer-wise semantic calibration \cite{CSAC_TKDE23}.

\section{\rfl}
\label{sec:rfl}

\subsection{Robustness Metrics}
\label{sec:rob_eval_metrics}
{\color{DarkRed}
While federated learning has emerged as a promising solution for many realistic settings, it often assumes that all clients are trustworthy and provide reliable information for the federation. However, due to its distributed nature, FL is vulnerable to various malicious manipulations by rogue clients. Existing \rfl{} strategies can be broadly categorized into two categories respectively towards \bayatt{} and \bacatt{}: \byzantoler{} (\cref{sec:byzantoler}) and \backdefen{} (\cref{sec:backdefen}). We provide the corresponding evaluation in the following.

\noindent$\bullet$~\textbf{\accdeclineimp{}: ${\bm{\mathcal{I}}}$}. The malicious attack goal for the \bayatt{} \cref{sec:byzantoler} is to hinder the federated performance compared with the accuracy of the global model ($A^{u}$), optimized in the benign setting without any attack.  We denote the ${\bm{\mathcal{A}}}^{u}_{Byz}$ as the performance under the \bayatt{}. We deﬁne the attack impact, ${\bm{\mathcal{I}}}$, to depict the accuracy drop as the following formulation:
\begin{equation}\small 
\setlength\abovedisplayskip{2pt} \setlength\belowdisplayskip{2pt}
{\bm{\mathcal{I}}} = {\bm{\mathcal{A}}}^{u} - {\bm{\mathcal{A}}}^{u}_{Byz}.
\label{eq:acc_degrade_impact}
\end{equation}

\noindent$\bullet$~\textbf{\attsuccerat{}: ${\bm{\mathcal{R}}}^{u}$}. For the \bacatt{} \cref{sec:backdefen}, we estimate the effectiveness by calculating the proportion of triggered samples classified as target labels. As illustrated in \cref{eq:backdoor}, we construct the trigger testing sample as $(\widetilde{x},\widetilde{y})$. Thus,  ${\bm{\mathcal{R}}}$ on the selected testing domain $u$ is formed as: 
 \begin{equation}\small 
\setlength\abovedisplayskip{2pt} \setlength\belowdisplayskip{2pt}
{\bm{\mathcal{R}}}^{u} = \frac{\sum (\arg \max(\widetilde{z})=\widetilde{y})}{|\widetilde{T}|}.
\label{eq:att_succ_rate}
\end{equation}

}

\subsection{\byzantoler{}} 
\label{sec:byzantoler}

To combat byzantine attackers, designing robust aggregation has become an effective paradigm. Existing solutions can be generally categorized into three classes:  
\begin{table}[t]\small
\caption{\textbf{Summary of essential characteristics for reviewed \byzantoler{} solutions} (\cref{sec:byzantoler}). }
\label{tab:summaryByzantineTolerance}
\vspace{-10pt}
\centering
{
\resizebox{\columnwidth}{!}{
\setlength\tabcolsep{1pt}
\renewcommand\arraystretch{1.2}
\begin{tabular}{rIc|c}
\hline\thickhline
\rowcolor{mygray}
Methods & Venue & Highlight \\
\hline\hline
\multicolumn{3}{l}{\textit{\distoler{}  \cref{sec:distoler}}} \\  
\multicolumn{3}{l}{\textbf{Limitation}: Fail in data heterogeneous federated learning
}\\ 
\hline
\multikrum{} \cite{Krum_NeurIPS17}  & \pub{NeurIPS'17} &  Average candidate gradient via \krum{} \\
\foolsgold{} \cite{FoolsGold_arXiv18}  & \pub{arXiv'18} & Filter sybils via contribution similarity \\
\dncagg{} \cite{DnC_NDSS21}  & \pub{NDSS'21} & SVD based spectral for outliers
detection  \\
\hline\hline
\multicolumn{3}{l}{\textit{\statoler{} \cref{sec:statoler}}} \\  
\multicolumn{3}{l}{\textbf{Limitation}: Require relative mathematical assumption}\\
\hline
\trimmedmedian{} \cite{Median_ICML18}  & \pub{ICML'18} & Remove via coordinatewise trimmed mean\\ 
\bulyan{} \cite{Bulyan_ICML18}  & \pub{ICML'18} & Agree each coordinate by major vectors\\ 
\rfa{} \cite{RFA_TSP22}  & \pub{TSP'22} & Geometric median and smoothed Weiszfeld\\ 
\hline\hline
\multicolumn{3}{l}{\textit{\protoler{} \cref{sec:protoler}}} \\
\multicolumn{3}{l}{\textbf{Limitation}:Require qualified dataset and homogeneity assumption}\\
\hline
\fltrust{} \cite{FLTrust_NDSS21}  & \pub{NDSS'21} & Allocate score via ReLU-clipped similarity\\ 
\sageflow{} \cite{Sageflow_NeurIPS21}  & \pub{NeurIPS'21} &   Entropy filtering and loss reweighting 
\end{tabular}}}
\vspace{-10pt}
\end{table}

\subsubsection{\distoler{}} 
\label{sec:distoler}
A group of robust aggregation strategy models normally compare the participating clients update differences and regard those significantly far from the overall direction as malicious clients and then exclude them from the aggregation process {\cite{Krum_NeurIPS17,FoolsGold_arXiv18,FABA_IJCAI19,AFA_arXiv19,DnC_NDSS21,MABRFL_IJCAI22}}. For example,  \multikrum{} \cite{Krum_NeurIPS17} selects the candidate gradient that is the closest to its neighboring clients. \foolsgold{} \cite{FoolsGold_arXiv18} leverages cosine similarity to identify malicious clients and allocate low weight. \faba{} \cite{FABA_IJCAI19} removes the outliers far from the mean value of the uploaded gradient. However, they basically rely on
the data homogeneity hypothesis, where each client shares the independent and identically distribution data distribution. Therefore, these methods are not applicable under data heterogeneous federated learning. 

\subsubsection{\statoler{}} 
\label{sec:statoler}
A handful of robust aggregation schemes focus on constructing diverse statistical criteria to select the circumvent the evil clients {\cite{Median_ICML18,Bulyan_ICML18,Attackadaptiveagg_ICJAIW21,RFA_TSP22}}. For instance, \rfa{} {\cite{RFA_TSP22}} calculates the geometric median with an alternating minimization function, which which empirically appears rapid convergence and can be interpreted as a numerically stable version of the Weiszfeld algorithm. Notably, \rfa{} gets rid of collecting the client individual contribution and is agnostic to the corruption level.
\bulyan{} \cite{Bulyan_ICML18} cooperates trimmed median \cite{Median_ICML18} to conduct a two-step meta-aggregation algorithm. Despite the certain advantages, they are also sensitive to the degree of data heterogeneity. They normally hypothesize that distributed data is constrained into a certain range. Besides, these methods are generally based on the statistics assumption, which presents a strong prior limitation for the realistic application.

\subsubsection{\protoler{}}
\label{sec:protoler}
One recent line of robust aggregation solutions {\cite{Sageflow_NeurIPS21,FLTrust_NDSS21,SmartFL_arXiv22}} leverages the proxy data to conduct the additional performance evaluation. The motivation behind this is that the proxy dataset normally shares a similar or constant semantic space with the private data and is assumed to be clean. Therefore, utilizing the proxy dataset for related evaluation modules is a reliable and convincing approach to effectively discriminate the benign and malicious ones. For example, \sageflow{} {\cite{Sageflow_NeurIPS21}} proposes an entropy-based filter and reweights aggregation based on empirical loss. \fltrust{} {\cite{FLTrust_NDSS21}} introduces ReLU-clipped cosine similarity and allocates high trust scores for those reliable clients. Notably, they depend on the auxiliary related data for examination, which hampers their practicability. Therefore, existing methods acquire strong assumptions for the data homogeneity or qualified related proxy dataset, as illustrated in \cref{tab:summaryByzantineTolerance}. 

\subsection{\backdefen{}} 
\label{sec:backdefen}
The potential backdoor attacks act as a serious threat to the federated learning system and have attracted a large number of interest in diverse defensive solutions to mitigate different backdoor attacks. Based on the distinct defensive kernel, the federated backdoor defenses can be classified into three major categories:

\subsubsection{\refdefen{}}
\label{sec:refdefen}
This type of efforts focuses on refining the aggregated global model to erase the possible backdoor attacks such as fine-tuning {\cite{FedPurning_arXiv20}}, logits distillation {\cite{FedMD_NeurIPS19,FedDF_NeurIPS20,FedRAD_ariv21,FCCL_CVPR22}}, and Bayesian learning {\cite{BNPFL_ICML19,SPAHM_NeurIPS19,FedMA_ICLR20,FEDBE_ICLR21}}. However, these strategies require a high-scale proxy dataset to tune the global model, which presents realistic deployment for the federated learning setting \cite{SmartFL_arXiv22}.  Besides, \refdefen{} fails to guarantee the backdoor erasing degree. Moreover, without the additional optimization regularization, it needs careful and complicated hyper-parameter configuration to optimize towards a satisfying condition to avoid the potential severe overfitting problem.

\subsubsection{\aggdefen{}}
\label{sec:aggdefen}
A group of solutions addresses this issue during the aggregation stage by excluding the malicious signals such as weights and gradients, from suspicious clients through anomaly behavior detection and dynamic weight allocation \cite{Krum_NeurIPS17,Median_ICML18,Bulyan_ICML18,RLR_AAAI21,DimKrum_EMNLP22}. Importantly, a part of these solutions are initially designed to defend against Byzantine attacks (as discussed in \cref{sec:byzantoler}) but have also proven to be effective in countering backdoor attacks. For example, \dimkrum{} \cite{DimKrum_EMNLP22} identifies abnormal and malicious client updates on a small fraction of dimensions with higher backdoor strengths, building on the \krum{} \cite{Krum_NeurIPS17}. \rlr{} \cite{RLR_AAAI21} adjusts the server learning rate through the sign information of client updates in terms of the dimension and round aspects. However, these solutions typically assume that each client distribution obeys the independent and identically distributed property, which does not hold for the heterogeneous federated learning setting {\cite{FLwithNonIID_arXiv18,ConvergenceofFedAvg_arXiv19,NonIIDQuagmireofFL_ICML20,RethinkingarchitectureinFL_CVPR22}}.

\subsubsection{\cerdefen{}}
\label{sec:cerdefen}
{\color{DarkRed}This family of methods aims to provide the certified robustness guarantee for each testing example, \ie, the prediction would not change even if some features in local training data of malicious clients have been modified within a certain constraint {\cite{PatchGuard_USENIX21,CertifiedviaRandomSmoothing_ICML19,DPA_ICLR21,CRFL_ICML21,ProvableFL_AAAI21,SparseFed_ICML22,FLIP_ICLR23}}. }Specifically, \provablefl{} \cite{ProvableFL_AAAI21} acquires multiple global models via optimizing on a randomly selected client subset. When inferring the testing example label, we take the majority vote results among the global models. \crfl{} \cite{CRFL_ICML21} controls the global smoothness via the clipping and smoothing parameter operations. \flip{} \cite{FLIP_ICLR23} combines trigger inversion techniques during the federated training process. Yet, it either requires the storage cost for a large number of models or brings performance degradation due to heavy parameter interpretation operation. Besides, some pre-define possible attack types and thus are not suitable for the open world.

\section{\ffl}
\label{sec:ffl}

\subsection{Fairness Metrics}
\label{sec:fair_eval_metrics}

{\color{DarkRed}
Federated learning is not a \textit{utopia}, which describes an imaginary community or society that holds highly desirable or near-perfect qualities for its members \cite{Utopian_DN03}.  Federated learning unavoidably brings the interest conflict problem. Specifically, the client spends both computation and communication costs to optimize and upload the federated models \cite{FLStrategies_arXiv16,MOCHA_NeurIPS17,Lotteryfl_arXiv20,Topology4CrossSiloFL_NeurIPS20,FedPara_ICLR22}. Besides, with respect to the quality and quantity of the data, the data value varies among different clients \cite{DAVINZ_ICML22,DataRightPricePri_Engineering23}. Their contributions to the federated system would not be suitable to regard equally. This federated learning requires incentive mechanisms to distinctly measure contribution and reward benefits for each client {\cite{FairandPrivacyFL_arXiv23,FAFL_TNNLS23}}. Existing \ffl{} strategies mainly deal with two types of unfairness: \rewaconf{} and \predbias{}, and thus naturally group into two methodology objective streams: \colfair{} (\cref{sec:colfair}) and \perfair{} (\cref{sec:perfair}). We provide the respective evaluation metrics as the following form.

\noindent$\bullet$~\textbf{\conmatchdeg{}: ${\bm{\mathcal{E}}}$}. Evaluating the client importance acts as an important step to reasonably allocate the profit \cite{CFFL_FL20,CGSV_NeurIPS21}. We argue that the reward mechanism is normally based on the pre-defined aggregation weight ($\alpha$ in \cref{eq:originfl}) to allocate reward. However, the realistic the client contribution could be could be directly reflected in the performance change. Specifically, we conduct the leave-one-out experiment and define the federated model without $i$ client as $w^{-i}\!=\!\frac{w-\alpha_i w_i}{1-\alpha_i}$. We measure the performance drop, $\Gamma_i$, if we remove the query client $i$. Then, we  calculate ${\bm{\mathcal{E}}}$ as: 
\begin{equation}\small 
\setlength\abovedisplayskip{2pt} \setlength\belowdisplayskip{2pt}
\begin{aligned}
\Gamma_i & = \overline{{\bm{\mathcal{A}}}} - \frac{1}{|{\mathcal{U}}|} \sum_{u \in {\mathcal{U}}}{\bm{\mathcal{A}}}^{u}_{-i},\\
\Gamma & = [\ldots,\frac{\Gamma_i}{\sum_{i \in k} \Gamma_i},\ldots]\in \mathbb{R}^{|K|},\\
{\bm{\mathcal{E}}} & = \frac{\Gamma \cdot \alpha}{||\Gamma||_2 ||\alpha||_2}.
\end{aligned}
\label{eq:contributionmatchdegree}
\end{equation}
The \cref{eq:contributionmatchdegree} reveals that a \colfair{} methodology would achieve a higher $\bm{\mathcal{E}}$ to reasonably allocate interest.

\noindent$\bullet$~\textbf{\perfovderv{}: ${\bm{\mathcal{V}}}$}. \perfair{} aims to maintain both high average accuracy and uniform accuracy distribution. Following \cite{qFFL_ICLR20,FedMGDA_TNSE20,FedFV_IJCAI21}, we utilize the standard deviation to measure the performance inconsistency of the selected algorithm across different testing distributions.
\begin{equation}\small 
\setlength\abovedisplayskip{2pt} \setlength\belowdisplayskip{2pt}
{\bm{\mathcal{V}}} =  \sqrt{\frac{1}{|{\mathcal{U}}|} \sum_{u \in {\mathcal{U}}} ({\bm{\mathcal{A}}}^{u} - \overline{{\bm{\mathcal{A}}}})^2} \%,
\label{eq:perform_var}
\end{equation}
where $\overline{{\bm{\mathcal{A}}}}$ denotes the average performance on the overall testing distributions. The larger the ${\bm{\mathcal{V}}}$ is, the more biased the fedeated performance is.
}
\subsection{\colfair{}}
\label{sec:colfair}
Contribution evaluation plays a crucial role in federated learning by assessing each client's contribution in a privacy-friendly manner \cite{ProtectAI_TDSC20,FairandPriFedModel_TPDS20}. Specifically, when one party contributes more, it may feel unfairly treated if all parties receive the same compensation regardless of their differing contributions. Hence, a fair contribution evaluation mechanism motivates clients to join the collaboration. Existing approaches basically follow the following ideologies.
\subsubsection{\indveval{}}
The individualized assessment paradigm focuses on measuring the client contribution based on individual contribution via the locally relevant information and the specific task performance. Specifically, the client-representative information normally builds on data collection cost {\cite{HFFL_arXiv20}}, contract theory {\cite{IncenDesignforEFLMobileNet_APWCS19,FLinVEC_Access20}}, Stackelberg Game {\cite{Stackelberg_JOTA73,FLwithStackelberg_NetworkLetters20}}, and computation bids {\cite{AuctionFL_TWC21}}. As for the specific task performance, some methods are based on the reputation mechanism to record the clients contributions \cite{CFFL_FL20,RRFL_ICMLW21,IncenDesignforEFLMobileNet_APWCS19,RFLforMobileNet_WC20,RFLforMobileNet_WC20,RepFLforSWN_IOTJ22,AuctionFL_TWC21}. For example, \cffl{} \cite{CFFL_FL20} evaluates the fairness by evaluating the validation performance on each client to represent the reputation metric. \rrfl{} \cite{RRFL_ICMLW21} considers the updates divergence between each local and shared global model to represent the reputation value.
Besides, several papers introduce different contribution metrics such as client bid and resource quality score \cite{AuctionFL_TWC21} and pair-wise mutual evaluation mechanism \cite{FairandPriFedModel_TPDS20}. However, \indveval{}  often assumes that both servers and clients are trustworthy, which can yield unreliable client contribution results in a semi-honest setting. Moreover, it may be less effective in dealing with significant data heterogeneity because clients with vastly different data distributions may exhibit noticeably different performances compared to others, resulting in the low reputation scores.

\subsubsection{\margeval{}}
As federated learning is a multip-party collaboration paradigm, some work evaluates the client marginal contribution to the federated performance. Notably, existing solutions normally take inspiration from the cooperative game theory \cite{GameTheory_97,KernelofCooperativeGame_65,CooperativeGame_12}, especially shapely value (SV) \cite{Shapley_97} to measure the player value. Related solutions focus on efficient shapely value measurement under the federated scenarios \cite{MeasureContriFL_BD19,EffandFairDataValuationforFL_FL20,CGSV_NeurIPS21,GTGShapleyFL_TIST22,FedCE_CVPR23}. For example, \cgsv{} \cite{CGSV_NeurIPS21} introduces the cosine gradient Shapley value to approximate the SV between the client and global updates. \fedce{} \cite{FedCE_CVPR23} estimates client contribution in both gradient and data space for medical image segmentation on the local evaluation.  However, these methods still face a heavy computation burden because the computational complexity of calculating SV is $\mathcal{O}(2^n)$. Furthermore, some require auxiliary dataset for performance validation, posing the additional data collection requirement.

\subsection{\perfair{}}
\label{sec:perfair}
The performance disparity means that the federated model presents a biased prediction preference, which appears in certain protected groups classification failure because of overfitting certain clients at the expense of other clients. Existing \perfair{} methods mainly focus on achieving uniform performance from the local optimization and parameter aggregation aspects \cite{AFL_ICML19,qFFL_ICLR20,SCAFFOLD_ICML20,SCAFFPD_arXiv23}. 

\subsubsection{\perdebioptim{}} 
\label{sec:perdebioptim}
A set of solutions proposes to mitigate the biased global performance by modifying the local objective function to satisfy the specific fairness constraints. One type focuses on the single worst client {\cite{AFL_ICML19,qFFL_ICLR20,FADE_SIGKDD21,GIFAIR_IJDS23}}.
For example, \afl{} \cite{AFL_ICML19} introduces the min-max approach to avoid overﬁtting to the particular client. However, naively focusing on the single worst objective can drag on the overall model utility. \qffl{} \cite{qFFL_ICLR20} calculates the min-max performance of all clients and penalizes those with larger empirical loss via the hyper-parameter. Another group leverages the multi-objective optimization \cite{FedMGDA_TNSE20,FCFL_NeurIPS21,Ditto_ICML21,FedBEAL_CVPR23,FedMDFG_AAAI23}. For instance, \fedmgda{} {\cite{FedMGDA_TNSE20}} designs a constraint multiple
gradient descent for a common descent direction.  \fcfl{} \cite{FCFL_NeurIPS21} utilizes the surrogate maximum function to consider the multi-objective optimization to achieve Pareto
optimality by controlling the gradient direction. However, these algorithms often assume that clients are benign and do not engage in malicious behavior against the federation. If a client maliciously reports its local empirical loss, it can lead to misleading updating directions \cite{FAFL_TNNLS23} and brings the federated optimization failure. 

\subsubsection{\perfairrewei{}} 
\label{sec:perfairrewei}
A mainstream of solutions has been developed to reweight the parameter aggregation to alleviate biased performance. These methods primarily rely on various parameter variance signals, such as gradients \cite{FedFV_IJCAI21}, predictive risk \cite{TERM_ICLR21,FedGA_CVPR23,FedCE_CVPR23,FairFed_AAAI23}. In particular, \fedfv{} \cite{FedFV_IJCAI21} utilizes the cosine similarity to measure gradient conflict and then modifies both magnitude and direction to avoid client conflict. \fedce{} \cite{FedCE_CVPR23} reduces the generalization gaps variance among different source domains to encourage optimization flatness and fairness. However, gradient estimation is based on the previous rounds and fails to synchronize with the latest updates. Additionally, prediction evaluation naturally involves additional validation datasets, which poses a serious data requirement for the method application.

\section{Setup}
\label{sec:setup}
\subsection{Experimental Datasets}
\label{sec:datasets}
According to the different data heterogeneity, we divide the existing benchmark datasets into the following two groups.
\subsubsection{\labelshift{} Datasets}
\label{sec:labelshift_datasets}
Existing research mainly utilizes the Dirichlet distribution: ${Dir}(\beta)$ to simulate the \labelshift{} distribution (\cref{eq:labelshift}) for experimental analysis {\cite{FedProx_MLSys2020,MOON_CVPR21,FedProc_arXiv21}}, where $\beta > 0$ is the concentration hyper-parameter to adjust the skewed level (imbalance degree of class composition). If $\beta$ is set to a smaller value, the local distribution appears more imbalanced from the global distribution. 

\noindent$\bullet$~\textbf{\cifarten{}} \cite{cifar_Toronto09} contains $50,000$ images for training and 10,000 images for the validation. Its image size is 32 $\times$ 32 within 10 categories.  

\noindent$\bullet$~\textbf{\cifarhun{}} \cite{cifar_Toronto09} is a famous image classification dataset, containing 32 $\times$ 32 images of 100 categories. Training and validating sets are composed of $50,\!000$  and $10,\!000$ images. 

\noindent$\bullet$~\textbf{\tyimg{}} \cite{ImageNet_CVPR09} is the subset of ImageNet with 100K images of size $64\times64$ with 200 classes scale.

\noindent$\bullet$~\textbf{\fashionmnist{}} \cite{fashionmnist_arXiv17} includes $70,000$ $28 \times 28$ grayscale fashion products pictures with with ten categories.

\subsubsection{\domainshift{} \& \outclishift{} Datasets}
\label{sec:domainandoutshift_datasets}
\begin{figure}[t]
\centering
\includegraphics[width=0.8\columnwidth]{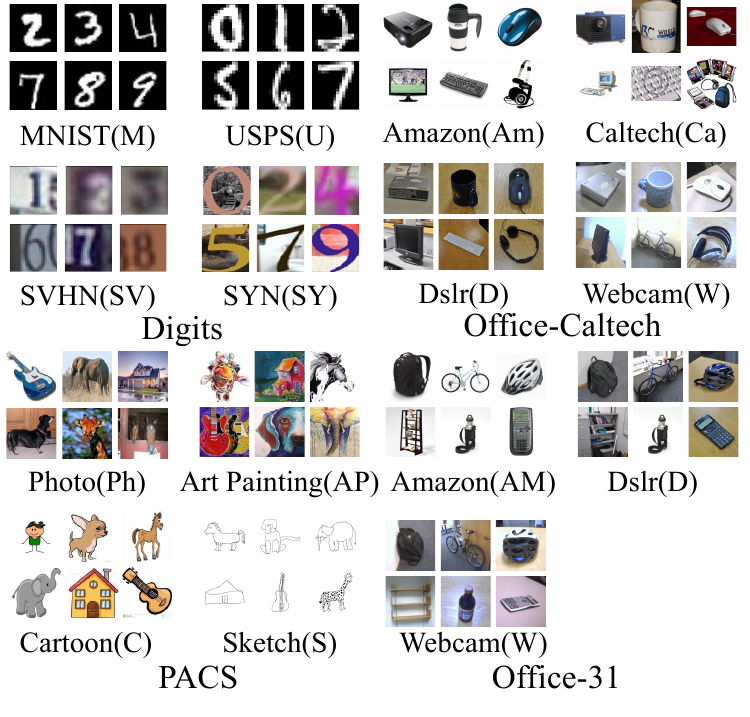}
\vspace{-10pt}
\caption{{Visualization} for \digits{} {\cite{MNIST_IEEE98,USPS_PAMI94,svhn_NeurIPS11,syn_arXiv18}}, \officecaltech{} {\cite{OffCaltech_CVPR12}},  \pacs{} \cite{PACS_ICCV17}, and \officeto{} \cite{office31_ECCV10}. Refer to \cref{sec:domainandoutshift_datasets}.}
\label{fig:visualization}
\vspace{-10pt}
\end{figure}

Both \domainshift{} and \outclishift{} settings consider the tasks that each distributed dataset is from a different domain with feature shift (\cref{eq:domainshift}). The crucial difference is that in the \outclishift{}, the evaluation solution is the leave-one-domain-out evaluation for all benchmarks, which means that randomly selecting one domain as the unseen client and all the left domains are used as source clients for collaboration. Furthermore, \cref{fig:visualization} shows example cases from relatively federated datasets.

\noindent$\bullet$~\textbf{\officecaltech{}} \cite{OffCaltech_CVPR12}  is built upon the Office dataset and Caltech256 {\cite{Caltech256_07}} datasets with 10 overlapping categories. It includes four diverse domains: \amazon{} (\amazonabbrv{}), \caltech{} (\caltechabbrv{}), \dslr{} (\dslrabbrv{}), and \webcam{} (\webcamabbrv{}).

\noindent$\bullet$~\textbf{\digits{}} is a numeral classiﬁcation task and includes four different domains: \mnist{} (\mnistabbrv{}) \cite{MNIST_IEEE98}, \usps{} (\uspsabbrv{})\cite{USPS_PAMI94}, \svhn{} (\svhnabbrv{}) \cite{svhn_NeurIPS11}, and \syn{} (\synabbrv{})\cite{syn_arXiv18} with ten categories.

\noindent$\bullet$~\textbf{\officeto{}} \cite{office31_ECCV10} has 31 classification number in three domains: \amazon, \dslr{}, and \webcam{}. The $31$ categories in the dataset consist of objects commonly encountered in office scenarios, such as laptops, keyboards, and, file cabinets.


\noindent$\bullet$~\textbf{\pacs{}} \cite{PACS_ICCV17} includes four domains: \photo{} (\photoabbrv{}) with $1,670$ images, \artpainting{} (\artpaintingabbrv{}) with 2,048 images, \cartoon{} (\cartoonabbrv{}) with 2,344 images and \sketch{} (\sketchabbrv{}) with 3,929 images. Each domain holds seven categories. 

\subsubsection{Data Augmentation}
\label{sec:data_augmentation}
We follow the wide-used data augmentation strategy in \cite{FedProc_arXiv21,MOON_CVPR21,FPL_CVPR23} to construct the local data augmentation strategy. We list these data augmentation introductions following the PyTorch notations:
\begin{fullitemize}
\item \texttt{RandomCrop}: The images are randomly cropped out with: $32 \times 32$, $224 \times 224$ for  respective federated scenarios.
\item \texttt{RandomHorizontalFlip}: The sample is horizontally flipped randomly with the probability $p=0.5$.
\item \texttt{Normalize}: Normalize the image with the mean value and standard deviation. 
\end{fullitemize}

\begin{table}[t]\small
\caption{Hyper-parameters chosen for different methods. Hyper-parameters in different methodologies may share the same notation but represent \textbf{distinct} meanings. See details in \cref{sec:implement_details}.}
\label{tab:chosen_hyper_param_table}
\vspace{-10pt}
\centering
{
\resizebox{\columnwidth}{!}{
\setlength\tabcolsep{1pt}
\renewcommand\arraystretch{1.2}
\begin{tabular}{rIl}
\hline\thickhline
\rowcolor{mygray}
Methods  & Hyper-Parameter \\
\hline\hline
\multicolumn{2}{l}{\gfl{} \cref{sec:gfl}} \\ 
\hline
\fedprox{} \cite{FedProx_MLSys2020}  
& Proximal weight $\mu:0.01$ \\
\scaffold{} \cite{SCAFFOLD_ICML20}  &  Global learning rate: $lr:0.25$\\
\fedproc{} \cite{FedProc_arXiv21}  
& Contrastive temp $\tau:1.0$ \\
\moon{} \cite{MOON_CVPR21}  
& Contrastive temp $\tau:0.5$; Proximal weight $\mu:1.0$ \\
\fedrs{} \cite{FedRS_KDD21} 
& Scaling factors $\alpha:0.5$ \\
\feddyn{} \cite{FedProc_arXiv21}  
& Proximal weight $\alpha:0.5$\\
\fedopt{} \cite{FedOPT_ICLR21}  
& Global Learning rate $\eta_g:0.5$ \\
\fedproto{} \cite{FedProto_AAAI22}  
& Proximal weight $\lambda:2$ \\
\fedlc{} \cite{FedLC_ICML22}  
& Scaling factor $\tau:0.5$ \\
\feddc{} \cite{FedDC_CVPR22} 
&  Penalized factor $\alpha:0.1$\\
\fedntd{} \cite{FedNTD_NeurIPS22}  
& Distill temp $\tau:1$; Reg weight $\beta:1$\\
\fpl{} \cite{FPL_CVPR23}  
&  Contrastive temp $\tau:0.02$ \\
\kdthreea{} \cite{KD3A_ICML21} 
& Confidence gate $g: [0.9,0.95]$ \\
\hline\hline
\multicolumn{2}{l}{\rfl{} \cref{sec:rfl}} \\ 
\hline
\multikrum{} \cite{Krum_NeurIPS17} 
& Evil ratio: $\Upsilon < 50 \%$; Top-K: $5$
\\

\bulyan{} \cite{Bulyan_ICML18} 
& Evil ratio: $\Upsilon < 50 \%$
\\

\trimmedmedian{}  \cite{Median_ICML18} 
& Evil ratio: $\Upsilon < 50 \%$
\\

\foolsgold{} \cite{FoolsGold_arXiv18} 
& Stability value $\epsilon=1e-5$ 
\\

\dncagg{} \cite{DnC_NDSS21} 
& Sub dimension $b:1000$ Filter ratio: $c:1.0$
\\

\fltrust{}  \cite{FLTrust_NDSS21} 
& Public epoch $E:20$
\\ 

\sageflow{}  \cite{Sageflow_NeurIPS21} 
& Threshold value $E_{th}:2.2$; Loss
exponent $\delta:5$
\\

\rfa{}  \cite{RFA_TSP22} 
& Iteration $E:3$
\\

\rlr{}  \cite{RLR_AAAI21} 
& Global learning rate $lr:1.0$; Robust threshold $\tau:4.0$
\\

\crfl{} \cite{CRFL_ICML21}
& Norm threshold $\rho:15$ Smooth level $\sigma:0.01$
\\
\hline\hline
\multicolumn{2}{l}{\ffl{} \cref{sec:ffl}} \\ 
\hline
\afl{} \cite{AFL_ICML19} & Regularization parameter $\gamma: 0.01$\\
\end{tabular}}}
\vspace{-10pt}
\end{table}


\begin{table}[t]\small
\caption{\textbf{Experiments Configuration of different federated scenarios}. Image Size is operated after the resize operation. $|C|$ denotes the classification scale. $|K|$ denotes the clients number. $E$ is the communication epochs for federation. $B$ means the training batch size} 
\label{tab:ExpeirmentsSetup}
\vspace{-10pt}
\centering
{
\resizebox{\columnwidth}{!}{
\setlength\tabcolsep{4pt}
\renewcommand\arraystretch{1.2}
\begin{tabular}{rIc|c|c|c|c|c|c}
\hline\thickhline
\rowcolor{mygray}
Scenario & Size & $|C|$ & Network $w$ & Rate $\eta$ & $|K|$ & $E$ & $B$\\
\hline\hline
\multicolumn{8}{l}{\textit{\labelshift{} Setting \cref{sec:labelshift_datasets}}} \\  
\hline
\cifarten{} & $32$  & $10$  & \simplecnn  & 1e-2 & 10 & 100 & $64$ \\ 
\fashionmnist{} & $32$  & $10$ & \simplecnn  & 1e-2 & 10 & 100 & $64$ \\ 
\mnist{} & $32$  & $10$ & \simplecnn  & 1e-2 & 10 & 100 & $64$\\ 
\cifarhun{} & $32$ & $100$  & \resnetfifty{} & 1e-1  & 10 & 100 & $64$\\ 
\tyimg{} & $32$ & $200$ & \resnetfifty{} & 1e-2  & 10 & 100 & $64$  \\ 
\hline\hline
\multicolumn{8}{l}{\textit{\domainshift{}} / \textit{\outclishift{} Settings} \cref{sec:domainandoutshift_datasets}} \\  
\hline
\digits{} & $32$  & $10$  &\resneteighteen{}  & 1e-2  & $4 / 3 $ & $50$ & $16$\\ 
\pacs{} & $224$ & $7$  &\resnetthirtyfour{}  & 1e-3 & $4 / 3 $ & $50$ & $16$\\ 
\officecaltech{} & $224$  & $10$  &\resnetthirtyfour{}  & 1e-3 & $4 / 3 $ & $50$ & $16$\\
\officehome{} & $224$ & $65$  &\resnetthirtyfour{}  & 1e-3 & $4 / 3 $ & $50$ & $16$\\ 
\end{tabular}}
}
\vspace{-10pt}
\end{table}

\begin{table*}[t]\small
\caption{{
\textbf{Quantitative \labelshift{} results} in term  of ${\bm{\mathcal{A}}}^{\mathcal{U}}$, ${\bm{\mathcal{A}}}^{u}$, and ${\bm{\mathcal{E}}}$ (\cref{eq:contributionmatchdegree}), on \cifarten{}, \cifarhun{}, \mnist{} and \fashionmnist{} scenarios. ${\bm{\mathcal{E}}}$ is evaluated under the $\beta=0.5$. 
Best in bold and second with underline. 
/ denotes that these methods are not available for these metrics, \ie, the value is zero or NaN.
 These notes also apply to the other tables. Please refer to \cref{sec:gen_eval_metrics} for metrics definition and \cref{sec:gen_compare} for experimental analysis. 
}}
\label{tab:quantiativ_label_shift}
\vspace{-10pt}
\centering
\scriptsize{
\resizebox{\linewidth}{!}{
\setlength\tabcolsep{2.pt}
\renewcommand\arraystretch{1.2}
\begin{tabular}{r||cccc|cIcccc|cIcccc|cIcccc|c}
\hline\thickhline
\rowcolor{lightgray} 
& 
\multicolumn{5}{cI}{\cifarten{}} & 
\multicolumn{5}{cI}{\cifarhun{}} & 
\multicolumn{5}{cI}{\mnist{}} & 
\multicolumn{5}{c}{\fashionmnist{}}
\\
\cline{2-21}  
\rowcolor{lightgray} \multirow{-2}{*}{Methods} 
& $1.0$ & $0.5$ & $0.3$ & $0.1$ 
& ${\bm{\mathcal{E}}}$
& $1.0$ & $0.5$ & $0.3$ & $0.1$   
& ${\bm{\mathcal{E}}}$
& $1.0$ & $0.5$ & $0.3$ & $0.1$ 
& ${\bm{\mathcal{E}}}$
& $1.0$ & $0.5$ & $0.3$ & $0.1$  
& ${\bm{\mathcal{E}}}$\\

\hline\hline

\fedavg{} \cite{FedAvg_AISTATS17}
& 70.64 & 66.96 & 63.92 & 60.43	& 0.354 
& 68.47 & 69.72 & 69.21 & 68.92	& 0.213 
& 99.44 & \underline{99.37} & 99.13 & 98.76	& 0.602
& 89.94 & 89.87 & 83.82 & \textbf{90.15}	& 0.462
\\

\fedprox{} \cite{FedProx_MLSys2020}
& \textbf{71.22} & 67.16 & 64.88 & 61.03	& 0.423 
& \textbf{72.37} & 70.19 & 63.48 &\underline{ 67.4}	& 0.773 
& 99.15 & \textbf{99.41} & \textbf{99.32} & 98.73	& 0.114 
& 89.87 & 89.97 & 88.69 & 83.57	& 0.524
\\

\scaffold{} \cite{SCAFFOLD_ICML20}
& 70.77 & \textbf{68.33} & \textbf{68.34} & 60.83	& /
& \underline{71.91} & \textbf{72.76} & \underline{69.82} &\textbf{68.24}	& /
& 99.41 & 99.12 & 98.95 & 96.95	& / 
& 89.83 & 89.73 & 88.32 & 81.27	& /
\\

\fednova{} \cite{FedNova_NeurIPS20}
& \underline{70.94} & 67.06 & 66.42 & \textbf{64.05}	& /
& 70.12 & 67.11 & 63.86 & 27.91	& / 
& 99.42 & 99.29 & 99.22 & \textbf{99.88}	& / 
& 90.20 & 89.81 & \textbf{89.03 }& \underline{84.39}	&/
\\

\moon{} \cite{MOON_CVPR21}
& 69.73 & \underline{68.07} & \underline{66.48} & 61.71	& 0.063 
& 71.47 & 69.51 & 69.09 & 65.53	& 0.412 
& \textbf{99.51} & 99.36 & 99.17 & 98.02	& 0.324
& \textbf{90.52} &\textbf{ 90.11} & \underline{88.95} & 82.92	& 0.614
\\

\fedrs{} \cite{FedRS_KDD21}
&70.14 & 66.036 & 63.89 & 59.47	& 0.184 
& 69.81 & 68.53 & 67.32 & 67.16	& 0.637 
& 99.34 & 99.33 & \underline{99.23} & \underline{98.93}	& 0.333 
& 90.01 & 89.40 & 88.47 & 77.54	& 0.579
\\

\feddyn{} \cite{FedDyn_ICLR21}
& 70.59 & 67.80 & 64.39 & 60.52	& 0.488
& 71.48 & \underline{71.25} & \textbf{70.28} & 66.81	& 0.583 
& \underline{99.48} & 99.31 & 99.10 & 98.71	& 0.059 
& \underline{90.24} & 89.97 & 88.59& 82.92	& 0.533
\\

\fedopt{} \cite{FedOPT_ICLR21}
& 70.44 & 66.70 & 65.95 & \underline{63.10}	& /
& 69.40 & 68.52 & 67.57 & 67.26	& / 
& 99.32 & \textbf{99.11} & 98.92 & 98.13	& /
& 90.06 & 89.65 & 88.79 & 83.41	& /
\\

\fedproto{} \cite{FedProto_AAAI22}
& 69.75 & 65.05& 56.45 & 48.74	& 0.319
& 70.07 & 70.83 & 68.32 & 67.36	& 0.759
& 99.44 & 99.26 & 99.12 & 98.69	& 0.323
& 90.17 & \underline{90.07} & 88.73 & 83.26	& 0.444
\\

\fedntd{} \cite{FedNTD_NeurIPS22}
& 51.43 & 35.06 & 37.37 & 22.18	& 0.647 
& 32.48 & 28.92 & 24.36 & 21.21	& 0.492
& 85.47 & 31.41 & 78.87 & 30.18	& 0.930
& 83.67 & 79.23 & 70.12 & 52.04 & 0.782




\end{tabular}}}
\vspace{-10pt}
\end{table*}


\begin{figure*}[t]
\label{fig:convergence_1}
  \vspace{-10pt}
\centering
\includegraphics[width=0.45\textwidth]{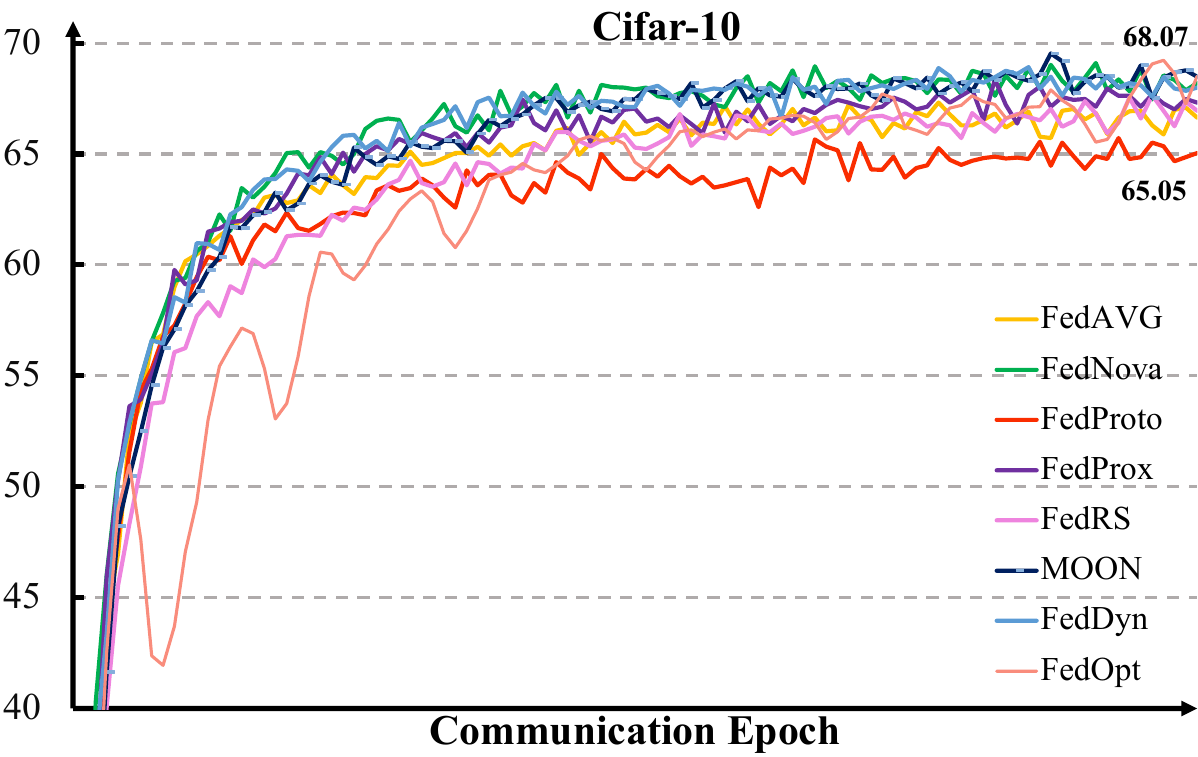}
\includegraphics[width=0.45\textwidth]{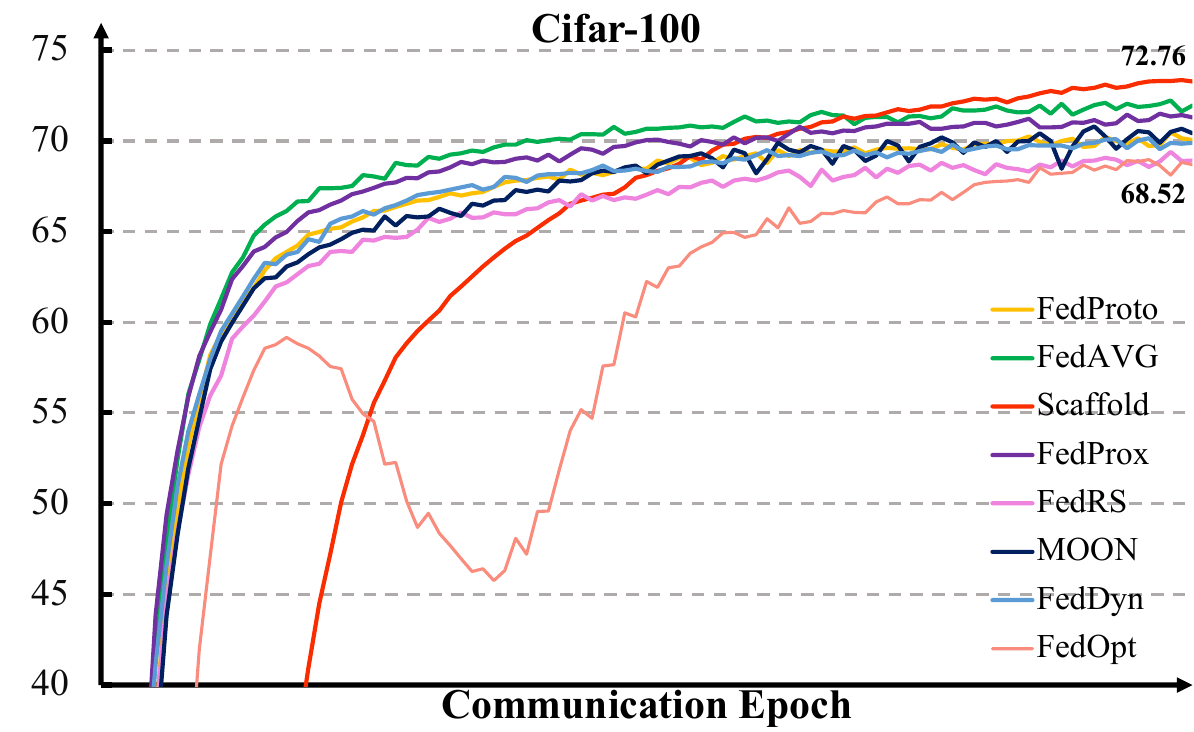}
  \vspace{-10pt}
\caption{Visualization of training curves of the test accuracy with Communication Epochs 100 with \cifarten{} and \cifarhun{} datasets ($\beta=0.5$).}
\vspace{-10pt}
\end{figure*}

\begin{figure*}[t]
\label{fig:convergence_2}
\centering
\includegraphics[width=0.45\textwidth]{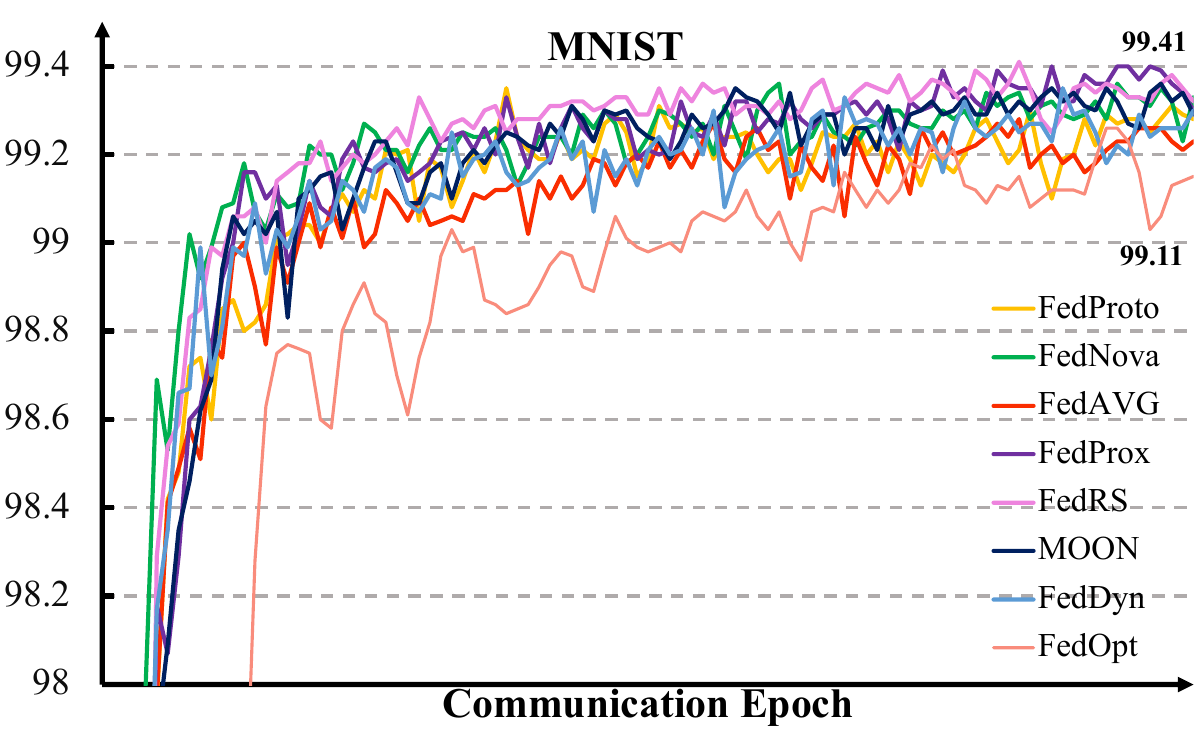}
\includegraphics[width=0.45\textwidth]{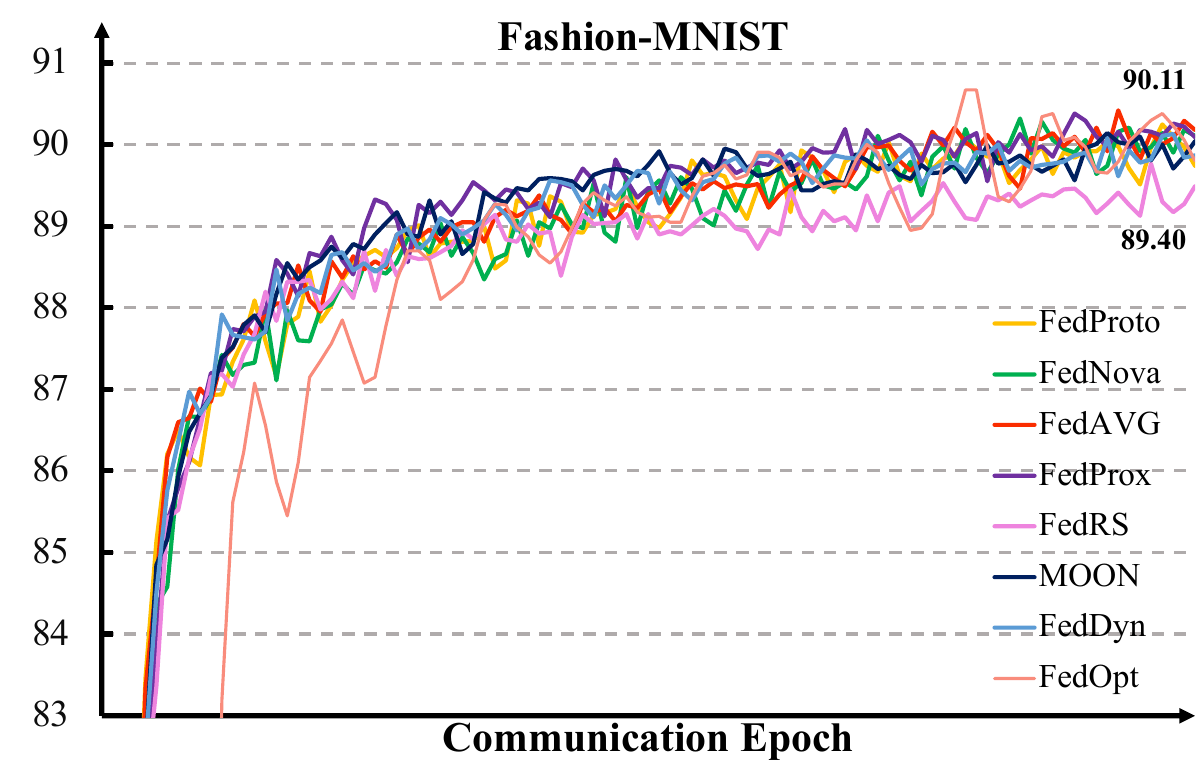}
\vspace{-10pt}
\caption{Visualization of training curves of the test accuracy
with Communication Epochs 100 with \mnist{} and \fashionmnist{} datasets ($\beta=0.5$).}
\vspace{-10pt}
\end{figure*}

\begin{table*}[t]\small
\caption{{
\textbf{Quantitative \domainshift{} results} in term  of ${\bm{\mathcal{A}}}^{\mathcal{U}}$, ${\bm{\mathcal{A}}}^{u}$, ${\bm{\mathcal{E}}}$ (\cref{eq:contributionmatchdegree}), and ${\bm{\mathcal{V}}}$ (\cref{eq:perform_var}) on \digits{}, \officecaltech{}, and \pacs{}. Refer to \cref{sec:gen_compare}.}}
\label{tab:quantiativ_domain_shift}
\vspace{-10pt}
\centering
\scriptsize{
\resizebox{\linewidth}{!}{
\setlength\tabcolsep{2.pt}
\renewcommand\arraystretch{1.1}
\begin{tabular}{r||cccc|cccIcccc|cccIcccc|ccc}
\hline\thickhline
\rowcolor{lightgray} 
& 
\multicolumn{7}{cI}{\digits{}} & 
\multicolumn{7}{cI}{\officecaltech{}} & 
\multicolumn{7}{c}{\pacs{}} 
\\
\cline{2-22}  
\rowcolor{lightgray} \multirow{-2}{*}{Methods} 
& \mnistabbrv{} & \uspsabbrv{} & \svhnabbrv{}  & \synabbrv{}
& ${\bm{\mathcal{A}}}^{\mathcal{U}}$ & ${\bm{\mathcal{E}}}$ & ${\bm{\mathcal{V}}}$
&  \amazonabbrv{} & \caltechabbrv{} & \dslrabbrv{} & \webcamabbrv{} 
& ${\bm{\mathcal{A}}}^{\mathcal{U}}$ & ${\bm{\mathcal{E}}}$ & ${\bm{\mathcal{V}}}$
& \photoabbrv{} & \artpaintingabbrv{} & \cartoonabbrv{} & \sketchabbrv{} 
& ${\bm{\mathcal{A}}}^{\mathcal{U}}$ & ${\bm{\mathcal{E}}}$ & ${\bm{\mathcal{V}}}$
\\

\hline\hline
\fedavg{} \cite{FedAvg_AISTATS17}
& 90.40 & 60.30 & 34.68& 46.99 & 58.09 & 0.024 & \underline{4.35}
& 81.99 & 73.21 & 79.37	& 67.93 & 75.62 & 0.653 & 0.379
& 76.09 & 64.19 & 83.50	& 89.40 & 78.30 & 0.279 & 0.911
\\

\fedprox{} \cite{FedProx_MLSys2020}
& 95.03 & 63.25 & 34.50	& 44.60 & 59.34 & 0.059 & 5.44
& 85.26 & 75.08 & 84.67 & 75.17 & \underline{80.23} & 0.717 & \textbf{0.273}
& 79.26 & 69.86 & 80.51	& 90.82 & \underline{80.19} & 0.170 & 0.612
\\

\scaffold{} \cite{SCAFFOLD_ICML20}
& 97.79 & 94.45 & 26.64	& 90.69 & \textbf{77.39} & / & 8.93
& 39.79 & 42.50 & 78.02	& 70.69 & 57.75 & / & \underline{0.281}
& 61.95  & 45.44 & 58.87	& 54.64 & 55.25 & / &\textbf{ 0.383}
\\

\moon{} \cite{MOON_CVPR21}
& 92.78 & 68.11 & 33.36& 39.28 & 58.36 & 0.287 & 5.72
& 84.42 & 75.98 & 84.67	& 68.97 & 78.51 & 0.678 & 0.539
& 74.44 & 64.19 & 83.92	& 89.17 & 77.93 & 0.321 & 0.924
\\

\feddyn{} \cite{FedDyn_ICLR21} 
& 88.91 & 60.34 & 34.57	& 50.72 & 58.65 & 0.161 & \textbf{4.06}
& 84.02 & 72.59 & 77.34	& 68.97 & 75.72 & 0.824 & 0.430
& 78.17 & 64.29 & 82.27	& 89.93 & 78.66 & 0.129 & 0.881
\\

\fedopt{} \cite{FedOPT_ICLR21}
& 92.71 & 87.62 & 31.32	& 87.92 & \underline{74.89} & / & 6.37 
& 79.05 & 71.96 & 89.34	& 74.48 & 78.71 & / & 0.480 
& 78.66 & 67.66 & 82.41	& 83.68 & 78.12 & / & \underline{0.410} 
\\

\fedproto{} \cite{FedProto_AAAI22} 
& 90.54 & 89.54 & 34.61	& 58.00 & 68.18 & 0.558 & 5.47 
& 87.79 & 75.98 & 90.0	& 79.31 &\textbf{83.27} & 0.556 & 0.410 
& 85.63 & 73.69 & 83.57	& 91.14 & \textbf{83.51} & 0.540 & 0.411 
\\



\fedntd{} \cite{FedNTD_NeurIPS22}
& 52.31 & 58.07 & 18.03	& 97.29 & 56.43 & 0.800 & 7.90
& 10.95 & 10.89 & 14.67	& 10.34 & 11.71 & 0.911 & 0.601
& 16.77 & 18.23 & 28.47	& 93.18 & 39.16 & 0.642 & 9.932
\\






\hline\hline
\multicolumn{10}{l}{Design for \perfair{} setting \cref{sec:perfair}} \\ 
\hline

\afl{} \cite{AFL_ICML19}
& 96.58 & 90.72 & 32.90 & 87.56 & 76.94 & 0.64 & 6.57
& 85.33 & 73.79 & 80.21 & 68.93 & 77.06 & 0.775 & 0.517
& 85.76 & 72.92 & 83.16 & 87.08 & 82.23 & 0.90 & 0.329
\\




\end{tabular}}}
\vspace{-10pt}
\end{table*}

\subsection{Implementation Details}
\label{sec:implement_details}
\noindent$\bullet$~\textbf{Training Setting}.
As for the uniform comparison evaluation, we follow {\cite{FedProc_arXiv21,MOON_CVPR21,FPL_CVPR23,DnC_NDSS21,Sageflow_NeurIPS21,FedCE_CVPR23,qFFL_ICLR20,CGSV_NeurIPS21}} and conduct the local updating round $U\!=\!10$. We use the SGD \cite{SGD_AoMS51} optimizer for all local updating optimization. The corresponding weight decay is $1e-5$ and momentum is $0.9$. The learning rate $\eta$ and communication epoch $E$ are different in various scenarios, as shown in \cref{tab:ExpeirmentsSetup}. Notably, the communication epoch is set according to when all federated approaches have little or no accuracy gain with more communication epochs. The local training batch size is $B\!=\!64$.  Furthermore, the \cref{tab:chosen_hyper_param_table} plots the chosen hyper-parameter for different methods.
We fix the random seed to ensure the experimental reproduction. Different methods are implemented on the PyTorch framework and trained on the NVIDIA GeForce RTX 3090.

\noindent$\bullet$~\textbf{Network Setting}. Following \cite{MOON_CVPR21,FedProc_arXiv21,FPL_CVPR23}, we utilize the CNN as the backbone for different scenarios, which has two $5 \times 5$ convolution layers with $2 \times 2$ max pooling and name is as \simplecnn{}. Besides, we utilize the \resnet{} network {\cite{ResNet_CVPR16}} family to support several large-scale dataset experiments.  We summarize the essential experiments configuration of different federated scenarios in \cref{tab:ExpeirmentsSetup}.

\noindent$\bullet$~\textbf{Malicious Setting}. For both byzantine and backdoor attacker scales, we set the malicious ratio $\Upsilon \!\in\! \{\! 0.2, 0.4 \! \}$ to represent the evils scale in the overall clients. Besides, for the Data-Based Byzantine Attack, the noise rate, $\epsilon$, is default set as $0.5$ for \pairflip{} and \symflip{} attacks.

\begin{table*}[t]\small
\caption{{
\textbf{Quantitative \outclishift{} results} in term of ${\bm{\mathcal{A}}}^{O}$ (\cref{eq:out_acc}) on \officecaltech{},  \digits{}, \pacs{}, \officeto{} scenarios. See details in \cref{sec:gen_compare}.}}
\label{tab:quantiativ_out_client__shift}
\vspace{-10pt}
\centering
\scriptsize{
\resizebox{\linewidth}{!}{
\setlength\tabcolsep{2.pt}
\renewcommand\arraystretch{1.2}
\begin{tabular}{r||cccc|cIcccc|cIcccc|cIccc|c}
\hline\thickhline
\rowcolor{lightgray} 
& 
\multicolumn{5}{cI}{\officecaltech{}} & 
\multicolumn{5}{cI}{\digits{}} & 
\multicolumn{5}{c}{\pacs{}} & 
\multicolumn{4}{c}{\officeto{}}
\\
\cline{2-20}  
\rowcolor{lightgray} \multirow{-2}{*}{Methods} 
& $\rightarrow$\caltechabbrv{}
& $\rightarrow$\amazonabbrv{} 
& $\rightarrow$\webcamabbrv{} 
& $\rightarrow$\dslrabbrv{} 
& AVG
& $\rightarrow$\mnistabbrv{} 
& $\rightarrow$\uspsabbrv{} 
& $\rightarrow$\svhnabbrv{}  
& $\rightarrow$\synabbrv{}
& AVG
& $\rightarrow$\photoabbrv{} 
& $\rightarrow$\artpaintingabbrv{} 
& $\rightarrow$\cartoonabbrv{} 
& $\rightarrow$\sketchabbrv{} 
& AVG
& $\rightarrow$\dslrabbrv{} 
& $\rightarrow$\amazonabbrv{} 
& $\rightarrow$\webcamabbrv{} 
& AVG
\\
\hline\hline

\fedavg{}   \cite{FedAvg_AISTATS17}
& 58.12 & 67.47 & 43.10	& 80.00 & 62.17
& 32.60 & 47.20 & 13.91	& 13.54 & 26.81
& 52.28 & 46.16 & 60.74	& 51.12 & 52.57
& 14.28 & 8.93 & 21.51 & 14.90
\\

\fedprox{}  \cite{FedProx_MLSys2020}
& 56.60 & 69.26 & 42.41	& 85.33 & 63.40
& 23.54 & 60.28 & 15.83	& 13.78 & 28.35
& 54.45 & 49.61 & 56.91	& 56.17 & 54.28
& 15.92 & 6.01 & 19.36 & 13.76
\\

\scaffold{}  \cite{SCAFFOLD_ICML20}
& 36.07 & 47.36 & 45.86	& 59.33 & 47.15
& 67.61 & 82.39 & 7.79	& 14.52 & \textbf{43.07}
& 43.85 & 23.81 & 45.07	& 39.79 & 38.12
& 12.44 & 5.58 & 10.88 & 9.63  
\\


\fedproc{}  \cite{FedProc_arXiv21}
& 47.41 & 60.84 & 42.41	& 66.66 & 54.33
& 24.34 & 43.37 & 10.15	& 13.09 & 22.73
& 56.94 & 30.95 & 56.02	& 49.94 & 48.46
& 19.39 & 4.91 & 10.38	& 11.56  
\\


\moon{}  \cite{MOON_CVPR21}
& 55.53 & 68.63 & 44.83	& 79.33 & 62.08
& 31.28 & 31.75 & 14.30	& 14.45 & 22.94
& 54.01 & 45.10 & 60.42	& 58.10 & \underline{54.40}
& 14.08 & 7.04 & 21.39 & 14.17  
\\

\feddyn{}  \cite{FedDyn_ICLR21}
& 59.99 & 66.42 & 40.34	& 81.99 & 62.18
& 28.74 & 56.08 & 14.36	& 11.88 & 27.76
& 51.40 & 43.19 & 60.57	& 50.71 & 51.46
& 14.08 & 7.86 & 17.85	& 13.26 
\\

\fedopt{}  \cite{FedOPT_ICLR21}
& 52.67 & 55.68 & 60.34	& 69.33 & 59.50
& 59.35 & 62.62 & 17.59	& 15.22 & \underline{38.69}
& 57.64 & 39.19 & 45.92	& 49.50 & 48.06
& 19.38 & 6.90	& 18.73 & 15.00
\\

\fedproto{}  \cite{FedProto_AAAI22}
& 60.35 & 66.94 & 58.62	& 76.00 & \textbf{65.47}
& 43.67 & 58.08 & 13.49	& 13.73 & 32.24
& 65.07 & 36.56 & 56.98	& 57.87 & 54.12
& 31.01 & 7.08 & 23.54	& \textbf{20.54}  
\\



\fedntd{}  \cite{FedNTD_NeurIPS22}
& 58.66 & 69.47 & 44.83	& 84.00 & \underline{64.23}
& 24.15 & 58.56 & 18.44	& 13.68 & 28.70
& 64.50 & 47.47 & 58.52	& 53.43 & \textbf{55.98}
& 17.75 & 7.12 & 27.97 & \underline{17.61}  
\\

\hline\hline
\multicolumn{10}{l}{Design for \fda{} setting \cref{sec:fda}} \\ 
\hline

\copa{}  \cite{COPA_ICCV21}
& 55.17 & 67.05 & 56.55 & 78.33 & 64.27
& 58.93 & 92.20 & 10.49	& 14.90 & 44.13
& 71.61 & 53.74 & 63.12	& 56.60 & 61.26
& 43.06 & 6.69 & 31.26	& 27.00
\\

\kdthreea{} \cite{KD3A_ICML21} 
& 54.73 & 70.00 & 68.61 & 75.33 & \textbf{67.16}
& 83.91 & 97.46 & 14.33 & 34.03 & \textbf{57.43}
& 76.99 & 56.91 & 67.63 & 55.70 & \textbf{64.30}
& 44.28 & 8.04 & 37.08 & \textbf{29.80}
\\

\hline\hline
\multicolumn{10}{l}{Design for \fdg{} setting \cref{sec:fdg}} \\ 
\hline
\copa{}  \cite{COPA_ICCV21}
& 57.32 & 66.31 & 48.27 & 70.00 & \textbf{60.47}
& 33.76 & 47.32 & 13.26 & 15.16 & 27.37
& 59.54 & 35.33 & 56.67 & 57.93 & \textbf{52.36}
& 21.22 & 5.48 & 19.49 & \textbf{15.39}
\\

\fedga{}  \cite{FedGA_CVPR23}
& 44.28 & 54.10 & 51.72	& 71.33 & 55.35
& 58.74 & 86.92 & 9.16	& 14.81 & \textbf{42.40}
& 59.00 & 35.01 & 43.20	& 53.60 & 47.70
& 22.24 & 5.15 & 10.63 & 12.67
\\

\end{tabular}}}
\vspace{-10pt}
\end{table*}

\begin{table*}[t]\small
\caption{{
\textbf{Quantitative \bayatt{} results} in term  of ${\bm{\mathcal{A}}}^{u}$, ${\bm{\mathcal{A}}}^{u}_{Byz}$, and ${\bm{\mathcal{I}}}$ (\cref{eq:acc_degrade_impact}) on \cifarten{}, \mnist{}, and \fashionmnist{} scenarios. \fltrust{} and \sageflow{}  utilizes \svhn{} as the proxy. The local optimization is  \fedprox{} \cite{FedProx_MLSys2020} with $\mu\!=\!0.01$. See \byzantoler{} comparison in \cref{sec:robust_compare}.}}
\label{tab:quantiativ_backdoor_attack}
\vspace{-10pt}
\centering
\scriptsize{
\resizebox{\linewidth}{!}{
\setlength\tabcolsep{1.pt}
\renewcommand\arraystretch{1.2}
\begin{tabular}{r||c|cc|c|ccIc|cc|c|ccIc|cc|c|ccIc|cc|c|cc}
\hline\thickhline
\cline{2-17}  
\rowcolor{lightgray} 
& \multicolumn{6}{cI}{\cifarten{}}  
& \multicolumn{6}{cI}{\fashionmnist{}}  
& \multicolumn{6}{cI}{\mnist{}}  
& \multicolumn{6}{c}{\usps{}}  
\\  
\cline{2-17}  
\rowcolor{lightgray}  
& \multicolumn{3}{c|}{$\beta\!=\!0.5$} 
& \multicolumn{3}{cI}{$\beta\!=\!0.3$ } 
& \multicolumn{3}{c|}{$\beta\!=\!0.5$ } 
& \multicolumn{3}{cI}{$\beta\!=\!0.3$} 
& \multicolumn{3}{c|}{$\beta\!=\!0.5$ } 
& \multicolumn{3}{cI}{$\beta\!=\!0.3$} 
& \multicolumn{3}{c|}{$\beta\!=\!0.5$ } 
& \multicolumn{3}{c}{$\beta\!=\!0.3$ } 

\\
\rowcolor{lightgray} 
& $\Upsilon\!=\!0.2$ & \multicolumn{2}{c|}{$\Upsilon\!=\!0.4$} 
& $\Upsilon\!=\!0.2$ & \multicolumn{2}{cI}{$\Upsilon\!=\!0.4$} 
& $\Upsilon\!=\!0.2$ & \multicolumn{2}{c|}{$\Upsilon\!=\!0.4$} 
& $\Upsilon\!=\!0.2$ & \multicolumn{2}{cI}{$\Upsilon\!=\!0.4$} 
& $\Upsilon\!=\!0.2$ & \multicolumn{2}{c|}{$\Upsilon\!=\!0.4$} 
& $\Upsilon\!=\!0.2$ & \multicolumn{2}{cI}{$\Upsilon\!=\!0.4$} 
& $\Upsilon\!=\!0.2$ & \multicolumn{2}{c|}{$\Upsilon\!=\!0.4$} 
& $\Upsilon\!=\!0.2$ & \multicolumn{2}{c}{$\Upsilon\!=\!0.4$} 
\\
\rowcolor{lightgray} \multirow{-4}{*}{Methods} 
& ${\bm{\mathcal{A}}}^{u}_{Byz}$ & ${\bm{\mathcal{A}}}^{u}_{Byz}$ & ${\bm{\mathcal{I}}}$
& ${\bm{\mathcal{A}}}^{u}_{Byz}$ & ${\bm{\mathcal{A}}}^{u}_{Byz}$ & ${\bm{\mathcal{I}}}$
& ${\bm{\mathcal{A}}}^{u}_{Byz}$ & ${\bm{\mathcal{A}}}^{u}_{Byz}$ & ${\bm{\mathcal{I}}}$
& ${\bm{\mathcal{A}}}^{u}_{Byz}$ & ${\bm{\mathcal{A}}}^{u}_{Byz}$ & ${\bm{\mathcal{I}}}$
& ${\bm{\mathcal{A}}}^{u}_{Byz}$ & ${\bm{\mathcal{A}}}^{u}_{Byz}$ & ${\bm{\mathcal{I}}}$
& ${\bm{\mathcal{A}}}^{u}_{Byz}$ & ${\bm{\mathcal{A}}}^{u}_{Byz}$ & ${\bm{\mathcal{I}}}$
& ${\bm{\mathcal{A}}}^{u}_{Byz}$ & ${\bm{\mathcal{A}}}^{u}_{Byz}$ & ${\bm{\mathcal{I}}}$
& ${\bm{\mathcal{A}}}^{u}_{Byz}$ & ${\bm{\mathcal{A}}}^{u}_{Byz}$ & ${\bm{\mathcal{I}}}$
\\
\hline\hline
\fedprox{}  \cite{FedProx_MLSys2020}
& \multicolumn{3}{c|}{ ${\bm{\mathcal{A}}}^{u}:$67.16}  	
& \multicolumn{3}{cI}{${\bm{\mathcal{A}}}^{u}:$64.88}  	

& \multicolumn{3}{c}{${\bm{\mathcal{A}}}^{u}:$89.97}  	
& \multicolumn{3}{cI}{${\bm{\mathcal{A}}}^{u}:$88.69}  

& \multicolumn{3}{c}{${\bm{\mathcal{A}}}^{u}:$99.41}  	
& \multicolumn{3}{cI}{${\bm{\mathcal{A}}}^{u}:$99.32}  

& \multicolumn{3}{c}{${\bm{\mathcal{A}}}^{u}:$96.70}  	
& \multicolumn{3}{c}{${\bm{\mathcal{A}}}^{u}:$96.69}  

\\
\hline
\multicolumn{10}{l}{\pairflip{} \cref{eq:pairflip}} 
\\ 
\hline

\multikrum{}  \cite{Krum_NeurIPS17}
& 50.21 & 46.85	& 20.31
& 46.99	& 43.91 & 20.82

& 82.20 & 47.59	& 42.38
& 80.79 & 82.51 &6.18

& 10.18 & 11.35 & 88.06
& 10.43	& 11.35 &  87.97

& 50.83 & 93.52 &3.18
& 93.41	& 51.11 & 45.58
\\

\bulyan{} \cite{Bulyan_ICML18}
& 46.88 & 44.06	& 20.68
& 10.00	& 10.00 & 54.88

& 82.62 & 80.76	&9.21
& 78.00 & 73.57 &15.12

& 97.01 & 98.18 & 1.23
& 93.21	& 92.13 &7.19

& 93.21 & 92.13 & 4.57
& 86.04	& 87.20 &9.49
\\

\trimmedmedian{}  \cite{Median_ICML18}
& 51.70 & 45.77	&21.39
& 19.94	& 10.67 &54.21

& 84.18 & 78.09	&11.88
& 81.76 & 77.89 &10.8

& 98.57 & 94.62 &4.79
& 93.25	& 92.90 &6.42

& 94.85 & 94.33 &2.37
& 91.72	& 92.05 &0.64
\\

\foolsgold{} \cite{FoolsGold_arXiv18}
& 60.09 & 56.80	&10.36
& 50.81	& 57.98 &6.90

& 86.97 & 86.07	&3.90
& 85.65 & 81.50 &7.19

& 97.25 & 97.80 &1.61
& 98.05	& 97.22 &2.10

& 77.69 & 91.77 &4.93
& 87.90	& 77.23 &19.46

\\

\dncagg{} \cite{DnC_NDSS21}
& \underline{62.67} & \underline{58.38} & \underline{8.78}
& \underline{60.41}	& \underline{59.96} & \underline{4.92}

& \underline{87.54} & \underline{87.76}	&\underline{2.21}
& \underline{87.22} & \underline{88.24} &\underline{0.45}

& \textbf{99.33} & \underline{99.07} &\underline{0.34}
& 98.85	& 98.70 &0.62

& \underline{95.94} & \underline{95.16} &\underline{1.54}
& 95.07	& 95.08 &1.61
\\

\fltrust{}  \cite{FLTrust_NDSS21}
& / & / & / 
& /	& / & / 

& / & / & / 
& /	& / & / 

& 11.35 & 11.35 &88.06
& 11.35 & 78.68	&20.64

& 13.15 & 13.15 &83.55
& 13.15	& 13.15 &83.54
\\

\sageflow{}  \cite{Sageflow_NeurIPS21}
& / & / & / 
& /	& / & / 

& / & / & / 
& /	& / & / 

& 99.28 & 99.03 &0.38
& \textbf{99.02} & \underline{98.73} &\underline{0.59}

& 95.36 & 94.34  &2.36
& \underline{96.15}	& \underline{95.37} &\underline{1.32}
\\

\rfa{}  \cite{RFA_TSP22}
& \textbf{66.84} & \textbf{66.31} &	\textbf{0.85}
& \textbf{62.28} & \textbf{61.54} &	\textbf{3.34}

& \textbf{89.67} & \textbf{89.73} & \textbf{0.24}
& \textbf{88.18} & \textbf{88.73} &\textbf{-0.04}

& \underline{99.12} & \textbf{99.10} &	\textbf{0.31}
& \underline{98.97} & \textbf{98.91} &	\textbf{0.41}

& \textbf{96.12} & \textbf{95.56} & \textbf{1.14}
& \textbf{96.30} 	& \textbf{96.08} & \textbf{0.61}
\\

\hline\hline
\multicolumn{10}{l}{\symflip{} \cref{eq:symflip}} \\ 
\hline

\multikrum{}  \cite{Krum_NeurIPS17}
& 52.18 & 46.48	&20.68
& 49.03 & 50.56	&14.32

& 81.87 & 85.52 &4.45
& 82.14 & 81.76 &6.93

& 10.02 & 91.76 &7.65
& 11.35	& 92.72 &6.60

& 81.20 & 93.06 &3.64
& 84.12	& \underline{93.79} & \underline{2.90}
\\

\bulyan{} \cite{Bulyan_ICML18}
& 50.73 & 38.38 &28.78
& 14.55 & 27.01 &37.87

& 84.15 & 82.15	&{7.82}
& 79.51 & 74.93 &13.76

& 97.16 & 97.52 &1.89
& 87.10	& 91.66 &7.66

& 91.46 & 89.71 &6.99
& 89.94	& 87.93 &8.76
\\

\trimmedmedian{}  \cite{Median_ICML18}
& 53.24 & 49.82 &	17.34
& 34.46 & 39.24 &	25.64

& 84.61 & 84.39 &	5.58
& 80.49 & 81.48 &7.21

& 98.50 & 98.08 & 1.33
& 92.16	& 96.25 &3.07

& 93.46 & 92.23 & 4.47
& 93.32	& 93.70 & 2.99
\\

\foolsgold{} \cite{FoolsGold_arXiv18}
& 61.37 & \underline{59.34} & \underline{7.82}
& 58.35 & 54.97 &	9.91

& 69.15 & 86.30 &3.67
& 82.34 & 84.27 &4.42

& 98.46 & 97.77 & 1.64
& 95.90	& 90.45 &8.87

& 83.02 & 78.07 &18.63
& 75.72 & 73.92 & 22.77
\\

\dncagg{} \cite{DnC_NDSS21}
& \underline{62.57} & 58.12 &	9.04
& \underline{61.94} & \underline{59.51} &	\underline{5.37}

& \underline{88.15} & \underline{87.23} &	\underline{12.74}
& \underline{86.33} & \textbf{87.83} & \textbf{0.86}

& \textbf{99.31} & \underline{98.99} & \underline{0.42}
& \underline{98.63}	& \underline{98.63} & \underline{0.69}

& \textbf{95.86} & \textbf{94.70} &	\textbf{2.00}
& 94.98 & 93.64 & 3.05
\\

\fltrust{}  \cite{FLTrust_NDSS21}
& / & / & / 
& /	& / & / 

& / & / & / 
& /	& / & / 

& 11.35 & 70.09 & 29.32
& 11.35	& 67.29 &32.03

& 60.41 & 52.83 & 43.87
& 59.31	& 13.15 &83.54
\\

\sageflow{}  \cite{Sageflow_NeurIPS21}
& / & / & / 
& /	& / & / 

& / & / & / 
& /	& / & / 

& 98.86 & 98.75 & 0.66
& 98.51	& 98.31 & 1.01

& 94.08 & 92.32 & 4.38
& \underline{95.33}	& 92.93 & 3.76
\\

\rfa{}  \cite{RFA_TSP22}
& \textbf{63.43} & \textbf{61.67} &	\textbf{5.49}
& \textbf{62.78} & \textbf{60.13} & \textbf{4.75}

& \textbf{89.44} & \textbf{88.30} & \textbf{11.67}
& \textbf{87.73} & \underline{87.49} &\underline{1.20}

& \underline{99.00} & \textbf{99.06} &	\textbf{0.35}
& \textbf{98.78} & \textbf{98.65} &\textbf{0.67}

& \underline{95.80} & \underline{94.57} & \underline{2.13}
& \textbf{95.98}	& \textbf{95.47} & \textbf{1.22}
\\

\hline\hline
\multicolumn{10}{l}{\randomnoise{} \cref{eq:randomnoise}}  \\ 
\hline

\multikrum{}  \cite{Krum_NeurIPS17}
& 10.00 & 13.06 & 54.1
& 29.25 & 14.11 & 50.77

& 10.00 & 21.71 &	68.26
& 75.55 & 25.60 & 63.09

& 11.35 & 13.42 &	85.99
& 11.35 & 21.04 & 78.28

& 89.25 & 15.07 &81.63
& 13.15 & 26.79 &69.90
\\

\bulyan{} \cite{Bulyan_ICML18}
& 51.04 & \underline{51.34} & \underline{15.82}
& 42.09 & 49.29 &	15.59

& 82.70 & \underline{87.24} & \underline{2.73}
& 81.70 & \underline{86.43} & \underline{2.26}

& 98.74 & 98.63 & 0.78
& 91.95	& \underline{98.32} & \underline{1.00}

& 94.27	& \underline{94.51} & \underline{2.19}
& \underline{92.59}	& \underline{95.34} & \underline{1.35}
\\

\trimmedmedian{}  \cite{Median_ICML18}
& 53.87 & 51.92 & 15.24
& 50.24 &  \underline{50.21} & \underline{14.67}

& 85.94 & 85.66 &	4.31
& 82.32 & 85.61 & 3.08

& 98.86 & \underline{98.85} & \underline{0.56}
& 94.36	& 98.18 & 1.14

& 94.80	& 13.15 &83.55
& \textbf{95.66}	& \textbf{95.59} & \textbf{1.10}
\\

\foolsgold{} \cite{FoolsGold_arXiv18}
& 50.01 & 32.85 &	34.31
& 49.60 & 27.45 &	37.43

& 85.98 & 35.82 &54.15
& 76.86 & 83.58 & 5.11

& 98.46 & 37.62 & 61.79
& 87.91	& 78.90 & 20.42

& 85.36	& 22.55 &74.15
& 54.10	& 55.92 &40.77
\\

\dncagg{} \cite{DnC_NDSS21}
& \textbf{59.64} & \textbf{56.95} &	\textbf{10.21}
& \textbf{60.00} & \textbf{56.45} & \textbf{8.43}

& \textbf{87.81} & \textbf{87.72} &	\textbf{2.25}
& \textbf{87.26} & \textbf{87.66} & \textbf{1.03}

& \textbf{99.31} & \textbf{98.97} & \textbf{0.44}
& \textbf{98.78} & \textbf{98.85} & \textbf{0.47}

& \textbf{95.73}	& \textbf{94.60} & \textbf{2.10}
& 95.31	& 94.28 &2.41
\\

\fltrust{}  \cite{FLTrust_NDSS21}
& / & / & / 
& /	& / & / 

& / & / & / 
& /	& / & / 

& 11.35 & 11.35 & 88.06
& 11.35	& 11.35 & 87.97

& 36.53	& 13.15 & 83.55
& 13.15	& 13.15 & 83.54
\\

\sageflow{}  \cite{Sageflow_NeurIPS21}
& / & / & / 
& /	& / & / 

& / & / & / 
& /	& / & / 

& 98.76 & 96.75 &2.66
& 93.14	& 89.85 &9.47

& 92.40	& 78.20 &18.50
& 86.02	& 75.63 &21.06
\\

\rfa{}  \cite{RFA_TSP22}
& \underline{56.37} & 10.64	&56.52
& \underline{55.88} & 15.45	&49.43

& \underline{87.11} & 64.10 &25.87
& \underline{85.32} & 72.30 &16.39

& \underline{99.15} & 95.40 &4.01
& \underline{98.26}	& 94.01 &5.31

& \underline{94.67}	& 67.49 &29.21
& 95.35	& 53.08 &43.61
\\

\hline\hline
\multicolumn{10}{l}{\minsum{} \cref{eq:minsumattack}} \\ 
\hline

\multikrum{}  \cite{Krum_NeurIPS17}
& 10.00 & 10.90	&56.26
& 42.20 & 10.02 &54.86

& 10.00 & 11.02 &78.95
& 80.78	& 10.00 &78.69

& 11.35 & 23.17 &76.24
& 10.43	& 11.35 &87.97

& 13.15 & 15.96 &80.74
& 13.15	& 13.15 &83.54
\\

\bulyan{} \cite{Bulyan_ICML18}
& 51.49 & 51.00	&16.16
& 42.99 & 40.07	&24.81

& 84.64 & \underline{85.84} & \underline{4.13}
& 80.23	& 84.21 &4.48

& 98.60 & 94.38 &5.03
& 92.40	& 90.14 &9.18

& 94.88 & 85.91 &10.79
& 92.91	& 93.36 &3.33
\\

\trimmedmedian{}  \cite{Median_ICML18}
& \underline{53.62} & \underline{53.71}	&\underline{13.45}
& 49.58 & \underline{51.76}	& \underline{13.12}

& 84.64 & 85.71 &4.26
& 83.24	& \underline{85.41} & \underline{3.28}

& 98.77 & \underline{98.76} & \underline{0.65}
& 96.80	& 92.90 &6.42

& \underline{95.12} & \textbf{95.75} & \textbf{0.95}
& 94.22	& \textbf{95.45} &\textbf{1.24}

\\

\foolsgold{} \cite{FoolsGold_arXiv18}
& 52.26 & 10.00 &57.16
& 47.83 & 10.00 &54.88

& 80.58 & 14.80 &75.17
& 80.20	& 19.36 &69.33

& 97.18 & 16.87 &82.54
& 98.71	& 97.22 &2.10

& 69.49 & 15.04 &81.66
& 64.16	& 13.12 &83.57

\\

\dncagg{}  \cite{DnC_NDSS21}
& \textbf{61.11} & \textbf{55.52} &	\textbf{11.84}
& \textbf{60.29} & \textbf{55.83} & \textbf{9.05}

& \textbf{87.63} & \textbf{87.80} & \textbf{2.17}
& \textbf{87.25} 	& \textbf{88.01} &\textbf{0.68}

& \textbf{99.19} & \textbf{99.20} &\textbf{0.21}
& \textbf{98.80} & \underline{98.70} &\underline{0.62}

& \textbf{95.34} & \underline{94.51} & \underline{2.19}
& \textbf{94.93} 	& \underline{95.35} & \underline{1.34}
\\

\fltrust{}  \cite{FLTrust_NDSS21}
& / & / & / 
& /	& / & / 

& / & / & / 
& /	& / & / 

& 61.57 & 12.99 &86.42
& 11.35 & 11.35 &87.97

& 13.15 & 15.04	&81.66
& 13.15 & 14.09 &82.60
\\

\sageflow{}  \cite{Sageflow_NeurIPS21}
& / & / & / 
& /	& / & /

& / & /  & /
& /	& /  & /

& 98.59 & 92.85 &6.56
& 92.30	& 85.01 & 14.31

& 87.07 & 14.09 &82.61
& 81.95	& 50.59 & 46.1
\\

\rfa{}  \cite{RFA_TSP22}
& 51.90 & 11.40	&55.76
& \textbf{60.29} & 14.22 &	50.66

& \underline{87.40} & 22.83 &67.14
& \underline{85.71}	& 61.18 &27.51

& \underline{99.05} & 94.39 &5.02
& \textbf{98.80}	& \textbf{98.91} & \textbf{0.41}

& 94.65 & 71.23 &25.47
& \textbf{94.93} 	& 57.83 &38.86
\\

\end{tabular}}}
\vspace{-10pt}
\end{table*}

\begin{table*}[t]\small
\caption{{
\textbf{Quantitative \bacatt{} results} in term  of ${\bm{\mathcal{A}}}^{u}$ and ${\bm{\mathcal{R}}}^{u}$ (\cref{eq:att_succ_rate}) on \cifarten{}, \mnist{},  and \usps{}. The local optimization algorithm is \fedavg{} \cite{FedAvg_AISTATS17}. We consider two types of backdoor attacks and abbreviate them as \backdoorabbrv{} {\cite{TargetBackdoor_arXiv17}} and \semanticbackdoorabbrv{} \cite{HowtoBackDoorFL_AISTATS20}. - means that these solutions are not applicable to these evaluations. Refer to \cref{sec:robust_compare} for \backdefen{} discussion.}}
\label{tab:quantiativ_bayzantine_attack}
\vspace{-10pt}
\centering
\scriptsize{
\resizebox{\linewidth}{!}{
\setlength\tabcolsep{1.pt}
\renewcommand\arraystretch{1.2}
\begin{tabular}{r||cccc|ccccIcccc|ccccIcccc|cccc}
\hline\thickhline
\cline{2-25}  
\rowcolor{lightgray} 
& \multicolumn{8}{cI}{\cifarten{}}  
& \multicolumn{8}{cI}{\mnist{}}  
& \multicolumn{8}{c}{\usps{}}  
\\  
\cline{2-25}  
\rowcolor{lightgray}  
& \multicolumn{4}{c|}{0.5} 
& \multicolumn{4}{cI}{0.3} 
& \multicolumn{4}{c|}{0.5} 
& \multicolumn{4}{cI}{0.3} 
& \multicolumn{4}{c|}{0.5} 
& \multicolumn{4}{c}{0.3}
\\
\cline{2-25}  
\rowcolor{lightgray}
& \multicolumn{2}{c|}{\backdoorabbrv}
& \multicolumn{2}{c|}{\semanticbackdoorabbrv}
& \multicolumn{2}{c|}{\backdoorabbrv}
& \multicolumn{2}{cI}{\semanticbackdoorabbrv}
& \multicolumn{2}{c|}{\backdoorabbrv}
& \multicolumn{2}{c|}{\semanticbackdoorabbrv}
& \multicolumn{2}{c|}{\backdoorabbrv}
& \multicolumn{2}{cI}{\semanticbackdoorabbrv}
& \multicolumn{2}{c|}{\backdoorabbrv}
& \multicolumn{2}{c|}{\semanticbackdoorabbrv}
& \multicolumn{2}{c|}{\backdoorabbrv}
& \multicolumn{2}{c}{\semanticbackdoorabbrv}
\\
\cline{2-25}  
\rowcolor{lightgray} 
\multirow{-4}{*}{Methods} 
& {${\bm{\mathcal{A}}}^{u}$} & ${\bm{\mathcal{R}}}^{u}$
& {${\bm{\mathcal{A}}}^{u}$} & ${\bm{\mathcal{R}}}^{u}$
& {${\bm{\mathcal{A}}}^{u}$} & ${\bm{\mathcal{R}}}^{u}$
& {${\bm{\mathcal{A}}}^{u}$} & ${\bm{\mathcal{R}}}^{u}$
& {${\bm{\mathcal{A}}}^{u}$} & ${\bm{\mathcal{R}}}^{u}$
& {${\bm{\mathcal{A}}}^{u}$} & ${\bm{\mathcal{R}}}^{u}$
& {${\bm{\mathcal{A}}}^{u}$} & ${\bm{\mathcal{R}}}^{u}$
& {${\bm{\mathcal{A}}}^{u}$} & ${\bm{\mathcal{R}}}^{u}$
& {${\bm{\mathcal{A}}}^{u}$} & ${\bm{\mathcal{R}}}^{u}$
& {${\bm{\mathcal{A}}}^{u}$} & ${\bm{\mathcal{R}}}^{u}$
& {${\bm{\mathcal{A}}}^{u}$} & ${\bm{\mathcal{R}}}^{u}$
& {${\bm{\mathcal{A}}}^{u}$} & ${\bm{\mathcal{R}}}^{u}$
\\
\hline\hline
\multicolumn{10}{l}{Focus on \byzantoler{}  \cref{sec:byzantoler}} \\ 
\hline


\bulyan{} \cite{Bulyan_ICML18}
& 47.61 & 28.73 & 44.61	& 17.12
& - & - & 11.12	& 19.56
& 96.95 & 14.77 & 92.13	& 0.45
& 87.70 & 11.13 & \textbf{87.86} & \textbf{0.10}
& 93.32 & 10.95 & 93.52 & 11.32
& 87.79 & 10.83	& 85.14 & 1.56
\\

\trimmedmedian{}  \cite{Median_ICML18}
& 51.34 & \textbf{22.49} & 52.21	& 13.70
& - & - & 14.78	& 51.66
& 98.07 & 99.18 & 98.44	& \underline{0.16}
& 96.65 & 89.42 & 96.72 & 0.61
& 94.62 & 71.52 & 94.24 & 4.82
& 92.05 & 84.17	& 94.77 & 2.40
\\

\foolsgold{} \cite{FoolsGold_arXiv18}
& \underline{60.69} & 62.54 & 60.50 & 13.06
& 58.58 & \underline{56.85}	& \underline{59.84} & 12.56
& 82.20 & 91.61 & 98.45	& 0.59
& 92.88 & 98.06 & 97.00 & 1.52
& 89.66 & 90.24 & 83.21 & 10.11
& 76.56 & 86.14	& 94.77 & 2.40
\\

\dncagg{} \cite{DnC_NDSS21}
& 59.30 & \underline{23.07} & \textbf{61.40} & 12.88
& \underline{60.03} & \textbf{42.79} & 59.80 & \textbf{9.76}
& \textbf{99.26} & \underline{10.39} & 99.13	& 0.20
& 98.53 & \underline{10.46} & \underline{98.79} & \underline{0.29}
& \underline{95.75} & \underline{9.62}  & \textbf{95.11} & \textbf{2.89}
& 96.14 & 16.89	& 94.86 & 1.81
\\

\fltrust{}  \cite{FLTrust_NDSS21}
& / & / & /	& /
& / & / & /	& /
& 95.31 & \textbf{8.71} & 97.84	& \textbf{0.00}
& 92.55 & \textbf{10.03} & 97.43 & 0.30
& 71.67 & 17.69 & 59.83 & 20.96
& \textbf{63.20} & \textbf{5.29} & 63.20 & 5.29 
\\

\sageflow{}  \cite{Sageflow_NeurIPS21}
& / & / & /	& /
& / & / & /	& /
& \underline{99.17} & 98.70 & \textbf{99.21}	& 0.53
& \underline{99.03} & 98.05 & 98.83 & 1.27
& 96.07 & 73.63 & 96.20 & 3.61
& 96.83 & 86.39	& 96.02 & 2.65
\\

\rfa{}  \cite{RFA_TSP22}
& \textbf{64.90} & 74.31 & \textbf{63.90} & 11.54
& \textbf{60.36} & 75.57 & \textbf{62.75} & 14.76
& 99.09 & 99.09 & \underline{99.12}	& 0.32
& \textbf{99.11} & 98.88 & 98.84 & 0.39
& \textbf{95.89} & \textbf{2.28} & \underline{95.75} & \underline{3.13}
& 97.04 & 39.59	& 95.89 & 2.28
\\

\hline\hline
\multicolumn{10}{l}{Focus on \backdefen{}  \cref{sec:backdefen}} \\ 
\hline

\rlr{}  \cite{RLR_AAAI21}
& 51.65 & 28.83 & 50.37	& \underline{10.60}
& - & -	& 44.80 & 20.74
& 94.77 & 10.54 & 93.11	& 0.40
& 91.11 & 22.69 & 92.94 & 0.35
& 89.20 & 10.78 & 92.00 & 12.65 
& \underline{87.00} & \underline{10.27}	& \underline{82.15} & \underline{1.44}
\\

\crfl{}  \cite{CRFL_ICML21}
& 59.27 & 63.29 & 58.59 & \textbf{9.52}
& 52.27	& 59.50 & 52.62 & \underline{11.66}
& 98.93 & 33.86	& 98.89 & 0.43
& 98.44 & 26.28	& 98.08 & 0.91
& 94.96 & 49.77	& 95.31 & 3.61
& 95.38 & 62.98	& \textbf{94.36} & \textbf{1.32}
\\

\end{tabular}}}
\vspace{-10pt}
\end{table*}



\section{Benchmark}
\label{sec:benchmark}
Next, we tabulate the performance of previously discussed solutions. For each of the reviewed ﬁelds, the most widely used dataset is selected for performance benchmarking. The performance scores are gathered from the reproduced results. As different methods are with different code bases and levels of optimization, it is hard to make completely fair comparison. Besides, for a small set of methods whose implementations are not well organized or publicly available, we directly borrow them from the unofficial code repository.

\subsection{Generalization Benchmark}
\label{sec:gen_compare} 

\subsubsection{Evaluation Metrics}
\croclienacc{} ${\bm{\mathcal{A}}}^{\mathcal{U}}$ is widely adopted in  the \croclishift{}: \labelshift{} and \domainshift{} settings. We further denote \outclienacc{} ${\bm{\mathcal{A}}}^{\mathcal{O}}$ under the \outclishift{} for generalizable performance evaluation.

\subsubsection{Results}
\cifarten{} \cite{cifar_Toronto09}, \cifarhun{} \cite{cifar_Toronto09}, \mnist{} \cite{MNIST_IEEE98}, and \fashionmnist{} \cite{fashionmnist_arXiv17} are arguably popular datasets for the \labelshift{} setting. The \cref{tab:quantiativ_label_shift} summarizes the results of ten methods on these four datasets. It reveals the first method \fedavg{} \cite{FedAvg_AISTATS17} proposed in 2017, to recent complicated solutions \cite{FedNTD_NeurIPS22}. For further comparison, we visualize the training curves of the testing evaluation accuracy along with the training process under the $\beta\!=\!0.5$.
Furthermore, as for the \domainshift{} scenario, we adopt the widely federated scenarios,\ie, \digits{} \cite{MNIST_IEEE98,USPS_PAMI94,svhn_NeurIPS11,syn_arXiv18}, \officecaltech{} \cite{OffCaltech_CVPR12}, and \pacs{} \cite{PACS_ICCV17}. As seen in \cref{tab:quantiativ_domain_shift}, \scaffold{} \cite{SCAFFOLD_ICML20} and \fedproto{} \cite{FedProto_AAAI22} present relative competitive performance.
Regarding the \outclishift{} task, we take into account the specialized \fda{} \cref{sec:fda} and \fdg{} \cref{sec:fdg} paradigms. Notably, \fda{} utilizes the unlabeled the unknown domain distribution during the training process and achieves the clear \outclienacc{}improvement. For example, \kdthreea{} achieves the $67.16$ accuracy on the \officecaltech{} scenario. 

\subsection{Robustness Benchmark}
\label{sec:robust_compare}

\subsubsection{Evaluation Metrics}
${\bm{\mathcal{A}}}^{u}_{Byz}$ means the testing accuracy under the \bayatt{}. And thus, \accdeclineimp{} ${\bm{\mathcal{I}}}$ denotes the decreased accuracy, compared with the benign federation results. Similarly, \attsuccerat{} ${\bm{\mathcal{R}}}^{u}$  evals the performance on the backdoor polluted datasets.

\subsubsection{Results}
The \cref{tab:quantiativ_bayzantine_attack} gathers the \bayatt{} results  for existing 
 \byzantoler{} methodologies.  We conduct the comparison under four widely adopted datasets, \ie, \cifarten{}, \fashionmnist{}, \mnist{}, and \usps{} datasets with both \databayatt{}: \pairflip{} and \symflip{}, and \modelbayatt{}: \randomnoise{} and \minsum. We select the popular \byzantoler{} methods from three types: \distoler, \statoler, \protoler{}. The results show that \dncagg{} reaches a relatively satisfying performance under different \bayatt{}.  \protoler{} appears the obvious disadvantage, additional related proxy dataset. The \cref{tab:quantiativ_backdoor_attack} evaluates the \bacatt{} results under two popular forms: \backdoorabbrv{}, and \semanticbackdoorabbrv{}. We further evaluate two \backdefen{}: \rlr{} \cite{RLR_AAAI21} and \crfl{} \cite{CRFL_ICML21}. It shows that \rfa{} and \crfl{} achieves satisfying defensive performance.
 
\subsection{Fairness Benchmark}
\label{sec:fairness_compare}

\subsubsection{Evaluation Metrics}
\conmatchdeg{}: ${\bm{\mathcal{E}}}$ and \perfovderv{} ${\bm{\mathcal{V}}}$   respectively evaluates the \colfair{} and \perfair{}.

\subsubsection{Results}
As shown in \cref{tab:quantiativ_label_shift} and \cref{tab:quantiativ_domain_shift}, few of the existing federated optimization takes the \colfair{} into federated objective account. Besides, \colfair{} also is largely impended under large local data distribution diversity such as the \domainshift{}. 
Regarding the \perfair{}, existing methods basically focus on minimizing the weighted empirical loss and thus bring the imbalanced performance. Notably,  global network utilization and server adaptive optimization seem to roundly alleviate the imbalanced performance on the multiple domains.

\subsection{Discussion}
\label{sec:summary}
From experiments, we draw several crucial conclusions. 

\noindent$\bullet$~\textbf{Reproducible Dilemma}. Across different federated fields, many methods do not describe the setup for the experimentation. Some of them even do not release the source code for implementation. Moreover,  different methods use various datasets and distinct backbone models. These make fair comparison impossible and hurt reproducibility.

\noindent$\bullet$~\textbf{Computation Cost}.
Another important fact discovered thanks to this study is the lack of information about execution time and memory cost. Federation can be basically divided into cross-device \cite{FLKeyboard_arXiv18} and cross-silo \cite{FLinMedicalApp_FDSE21} settings \cite{FederatedMLConApp_TIST19}. Specifically, many methods claim that they greatly improve federated performance improvement. However, they properly utilize the heavy memory cost or tedious execution time. This void is due to the fact that most solutions focus only on accuracy metrics without any concern about running time efﬁciency or memory requirements. However, computation-efficient acts as an important role in the federated realistic setting and should be given high-level priority.

\noindent$\bullet$~\textbf{Thinking Everything Failure}. 
Existing federated methods primarily focus on addressing separate problems, such as data heterogeneity \cite{NonIIDFL_FGCS22,FLONNonIID_TKDE22}, malicious attacks \cite{SurveyonSecurityandPrivacyofFL_FGCS21,ByzantineAttacksinFL_TrustCom22}, and others. Although these methods often achieve satisfactory performance on specific evaluation metrics, we are curious whether a feasible federated paradigm exists to simultaneously handle various aspects, like a unified federation.






\section{Outlook}
\label{sec:outlook}

\subsection{Future Direction}
\label{sec:future_direction}
Based on the reviewed research, we list several future research directions that we believe should be pursued.

\noindent$\bullet$~\textbf{Generalization and Robustness Dilemma}.
Distributed data inherently exhibits the heterogeneous property and generalization aims to incorporate diverse client knowledge. Concurrently, Robustness aims to identify and mitigate the malicious clients effect on the federated system \cite{RiFT_ICCV23,FedIPR_PAMI22}. However, data heterogeneity in federated learning brings that certain local distributions deviate significantly from the primary trend and may be erroneously categorized as malicious ones. This misclassification restricts the effectiveness of federated performance because some benign clients are rejected from the federation.  

\noindent$\bullet$~\textbf{Generalization and Fairness Trade-Off}. Generalization primarily focuses on fitting the distributed data distribution to enhance the mean accuracy ${\bm{\mathcal{A}}}^{\mathcal{U}}$ on the test distribution collection $\mathcal{U}$. However, \ffl{}, especially \perfair{}, places emphasis on the client-wise or group-wise uniform performance and requires less performance variance on different testing scenarios. Thus, considering both generalization and robustness, it naturally involves a multi-task optimization: striving for both higher overall performance and uniform individual testing distribution performance. Another Prisoner dilemma is that generalization highlights the federated convergence speed, which normally achieves via fitting on the major client distribution at the cost of ignoring some minority groups. However, this optimization objective presents a contradictory direction with fairness, which requires uniform performance on different testing distributions.

\noindent$\bullet$~\textbf{Robustness and Fairness Cooperation}.
Fairness aims to ensure long-term and stable multi-party federated cooperation. Robustness focuses on eliminating the malicious influence on the federation. Thus, to some extent, they promote each other. Specifically, achieving \colfair{} requires accurate client contribution measurement, which helps detect those malicious ones with meaningless contributions. On the flip side, with reliable robustness, only benign ones are enrolled in the federated system, which boosts the better interest allocation.

{
\noindent$\bullet$~\textbf{Vertical FL meets Generalization, Robustness, Fairness}.
\vfl{} (VFL) has attracted attention in federated learning, and thus the generalization, robustness, and fairness problems in VFL have become burgeoning research directions \cite{VFL_arXiv22, VFLChallMethoExper_arXiv22}.
{\color{DarkRed}With regard to the generalization aspect, a primary challenge in VFL stems from the need to align and integrate diverse feature sets pertaining to identical entities across varied domains {\cite{VFLAuto_JMLR21,VFLUnsupervised_TBD22}}. Moreover, several domains may have missing features for certain entities, which complicates the learning process as it is not straightforward to handle such incomplete information without biasing the model. The questions for aligning features and conducting de-bias training are yet to be resolved.}
As for the robustness, moving beyond the Byzantine and Backdoor attacks in robust FL,  VFL should address data inference assaults, wherein malicious clients endeavor to deduce privacy information from the model updates or the final model. Such attempts encompass label inference and feature inference attacks \cite{vfl_attack_TBD22, label_infer_usenix22,FeatInferVFL_ICDE21,Cafe_NeurIPS21,VFLUnsupervised_TBD22}. 
There is a need for the development of more defense strategies against them, \eg, secure aggregation protocols, robust model architectures, and decentralized trust frameworks) which should not only aim to safeguard data but also strive to maintain a balance between model performance and computational efficiency.
{\color{DarkRed}Regarding the fairness concept, the imbalance in contribution fairness happens when one client data contributes more significantly to the predictive power of the model than others \cite{VFL_fairness_arxiv21, FairVFL_NeurIPS22,VFPS_NeurIPS22,LESSVFL_ICML23}. Identifying fairness-sensitive features among participants is crucial, and the development of collaborative, bias-mitigating algorithms is necessary for fair representation.}
}

\noindent$\bullet$~\textbf{Federated Learning with Large Language Model}. Recently, the \llm{} (\llmabbrv{}) represents a significant breakthrough, \eg, GPT family {\cite{GPT2_OPENAI19,GPT3_NeurIPS20,GPT4_arXiv23}}, PaLM series {\cite{PaLM_arXiv22,PaLMv2_arXiv23}}. However,  \llmabbrv{} requires the high-quality dataset to improve performance, which incurs expensive data collection cost. Notably, federated learning presents a  viable solution by establishing collaboration to eliminate the large-scale data scrape requirement \cite{FoudationmeetsFL_arXiv23}. 
However, a major hindrance derives from the communication cost. Specifically, \llmabbrv{} normally contains the enormous parameters scale and presents the strict challenge on the federation communication burden, which relies on the parameter exchange for multiple-party collaboration. Besides, directly sharing overall parameters largely threatens the company interests. Therefore, considering both the communication cost and intellectual property\cite{IPR_06}, achieving controllable and privacy-preserving communication media is important for the federated \llm{} deployment.

\subsection{Conclusion}
\label{sec:conclusion}
To our knowledge, this is the ﬁrst survey to comprehensively review recent progress in federated learning from generalization, robustness, and fairness aspects. 
We provided the reader with the necessary background knowledge and summarized more than 100 federated methods according to various criteria, including task settings, learning strategies, and technique contributions.
We also present benchmarking results of 8 widely used federated datasets.
We discuss the results and provide insight into the shape of future research directions and open problems in the ﬁeld.
In conclusion, federated learning has achieved notable progress attributed to the striking research development, but several challenges still lie ahead.

\ifCLASSOPTIONcaptionsoff
  \newpage
\fi



%
	\vspace{-4pt}
{\small
\bibliographystyle{IEEEtran}

\bibliography{egbib}
}

\vfill


\end{document}